\documentclass{article} % For LaTeX2e

\usepackage{main,times}
\iclrfinalcopy

\usepackage{amsmath,amsfonts,bm}

\def\eqref#1{equation~\ref{#1}}
\def\1{\bm{1}}

\DeclareMathAlphabet{\mathsfit}{\encodingdefault}{\sfdefault}{m}{sl}
\SetMathAlphabet{\mathsfit}{bold}{\encodingdefault}{\sfdefault}{bx}{n}

\usepackage{hyperref}
\usepackage{url}

\usepackage{hyperref}
\usepackage{cleveref}
\usepackage{verbatim}

\usepackage{wrapfig}  % Add this to your preamble
\usepackage{graphicx} 
\usepackage{subcaption}
\usepackage{listings}
\usepackage{algorithm}
\usepackage{enumitem}

\usepackage{subcaption} % 
\usepackage[toc,page,header]{appendix}

\usepackage{bbding}
\usepackage{graphicx}
\usepackage{booktabs}
\usepackage{makecell}
\usepackage{multirow}
\usepackage{tabularx}
\usepackage{float}
\usepackage{xcolor}
\usepackage{subcaption}
\usepackage{threeparttable}
\usepackage{caption}

\usepackage[table]{xcolor} 
\definecolor{myblue}{RGB}{220,230,241}

\usepackage[affil-it]{authblk}

\author[1]{Baode Wang$^{*}$}
\author[2]{Biao Wu$^{*}$}
\author[1]{Weizhen Li$^{*}$}
\author[3]{Meng Fang$^{*}$}
\author[1]{Zuming Huang$^{\dagger}$}
\author[1]{Jun Huang}
\author[1]{Yanjie Liang}
\author[1]{Haozhe Wang}
\author[2]{Ling Chen}
\author[1]{Wei Chu}
\author[1]{Yuan Qi$^{\ddagger}$}

\affil[1]{INFLY Tech}
\affil[2]{Australian Artificial Intelligence Institute}
\affil[3]{University of Liverpool}

\usepackage{minitoc}

\title{Infinity-Parser: Layout-Aware Reinforcement Learning for Scanned Document Parsing}

\begin{document}

\maketitle
\begingroup
\renewcommand{\thefootnote}{\fnsymbol{footnote}}
% \footnotetext[1]{Equal contribution.}
\footnotetext[2]{Lead contributor. Contact: \texttt{zuminghuang@gmail.com}}
\footnotetext[3]{Corresponding Author.}
\endgroup

\begin{abstract}
% Automated parsing of scanned documents into richly structured, machine‑readable formats remains a critical bottleneck in Document AI, as traditional multi‑stage pipelines suffer from error propagation and limited adaptability to diverse layouts. We introduce layoutRL, an end‑to‑end reinforcement‑learning framework that trains models to be explicitly layout‑aware by optimizing a composite reward of normalized edit‑distance, paragraph‑count accuracy, and reading‑order preservation. Leveraging our newly released dataset—Infinity‑Doc‑400K, which combines 400K high‑fidelity synthetic scanned document parsing data with expert‑filtered real‑world documents—we instantiate layoutRL in a vision‑language‑model–based parser called Infinity‑Parser. Evaluated on English and Chinese benchmarks for OCR, table and formula extraction, and reading‑order detection, Infinity‑Parser achieves new state‑of‑the‑art performance in both accuracy and structural fidelity, outpacing specialist pipelines and general‑purpose vision‑language models. We will publicly release our code and dataset to accelerate progress in robust document understanding.

Document parsing from scanned images into structured formats remains a significant challenge due to its complexly intertwined elements such as text paragraphs, figures, formulas, and tables. Existing supervised fine-tuning methods often struggle to generalize across diverse document types, leading to poor performance, particularly on out-of-distribution data. This issue is further exacerbated by the limited availability of high-quality training data for layout-aware parsing tasks. To address these challenges, we introduce LayoutRL, a reinforcement learning framework that optimizes layout understanding through composite rewards integrating normalized edit distance, paragraph count accuracy, and reading order preservation. To support this training, we construct the Infinity-Doc-400K dataset, which we use to train Infinity-Parser, a vision-language model demonstrating robust generalization across various domains. Extensive evaluations on benchmarks including OmniDocBench, olmOCR-Bench, PubTabNet, and FinTabNet show that Infinity-Parser consistently achieves state-of-the-art performance across a broad range of document types, languages, and structural complexities, substantially outperforming both specialized document parsing systems and general-purpose vision-language models. We will release our code, dataset, and model to facilitate reproducible research in document parsing.

\end{abstract}

\section{Introduction}

Document parsing aims to convert scanned documents into structured, machine-readable formats and represents one of the core tasks in document intelligence~\citep{DocumentParsing1,MinerU,got2,xia2024docgenome,zhang2024document}. Unlike traditional OCR that focuses solely on text recognition, document parsing requires comprehensive recovery of hierarchical document structures, including the dependency relationships among elements such as paragraphs, headers, tables, and formulas—a capability that is crucial for downstream applications including legal contract analysis, scientific literature mining, and financial report processing. Traditional approaches typically rely on multi-stage pipelines that decompose the task into supervised sub-tasks—such as layout detection, OCR, table recognition, and formula recognition—followed by heuristic post-processing to reconstruct document structure~\citep{Nougat,Vary,fox,got2,Qwen-VL,InternVL}. However, such pipeline-based methods are prone to error propagation and exhibit limited adaptability when confronted with diverse layout variations.

Recent approaches primarily reformulate document parsing as end-to-end perception tasks using vision-language models (VLMs) trained through supervised fine-tuning (SFT). However, this paradigm faces fundamental limitations. Although SFT provides token-level supervision, it often overfits to surface patterns rather than learning generalizable structural representations. This limitation is further compounded by the scarcity of large-scale, high-quality training data for document parsing, which hinders models from acquiring layout-aware knowledge. To overcome these limitations, reinforcement learning (RL) presents a promising alternative, having demonstrated strong generalization capabilities in vision and multimodal tasks, where outcome-based rewards help models learn transferable representations~\citep{huang2025visionr1,liu2025visionreasoner,wang2025vl}. This raises a fundamental question: can reinforcement learning drive models toward generalizable layout parsing rules? Unfortunately, current RL approaches remain limited by coarse-grained, binary outcome rewards that fail to provide the fine-grained, layout-aware supervision necessary for modeling complex document layouts~\citep{guo2025deepseek,shao2024deepseekmath}. Therefore, how to effectively apply RL to document parsing remains an underexplored yet critical challenge.

% However, this paradigm faces fundamental limitations: while SFT provides robust token-level supervision, it tends to memorize surface patterns rather than learn generalizable structural understanding, resulting in poor structural consistency and limited cross-domain generalization in complex, diverse document scenarios~\citep{chu2025sft}. 

% his raises a fundamental question: in document parsing, does RL can drive models toward generalizable layout parsing rules? However, current RL approaches are still limited by coarse-grained, binary outcome rewards, which fail to provide the fine-grained, layout-aware supervision needed for modeling complex document layouts~\citep{li2023rlhf,zheng2025easyr1}. Designing effective layout-aware reward functions tailored for document parsing remains an underexplored challenge.

\begin{figure}[t!]
  \centering
  % 左图
  \includegraphics[width=0.48\linewidth]{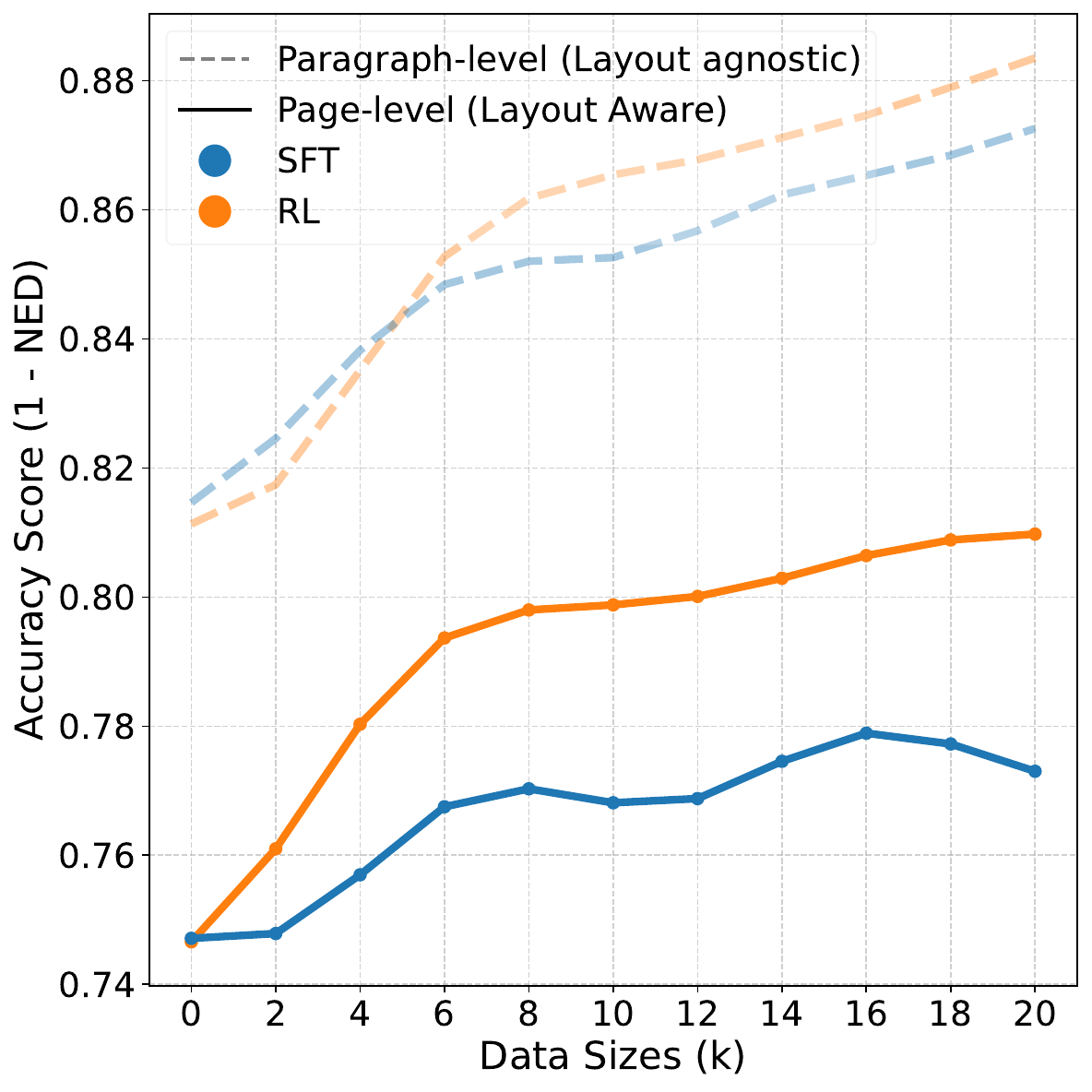}
  \hfill
  \includegraphics[width=0.48\linewidth]{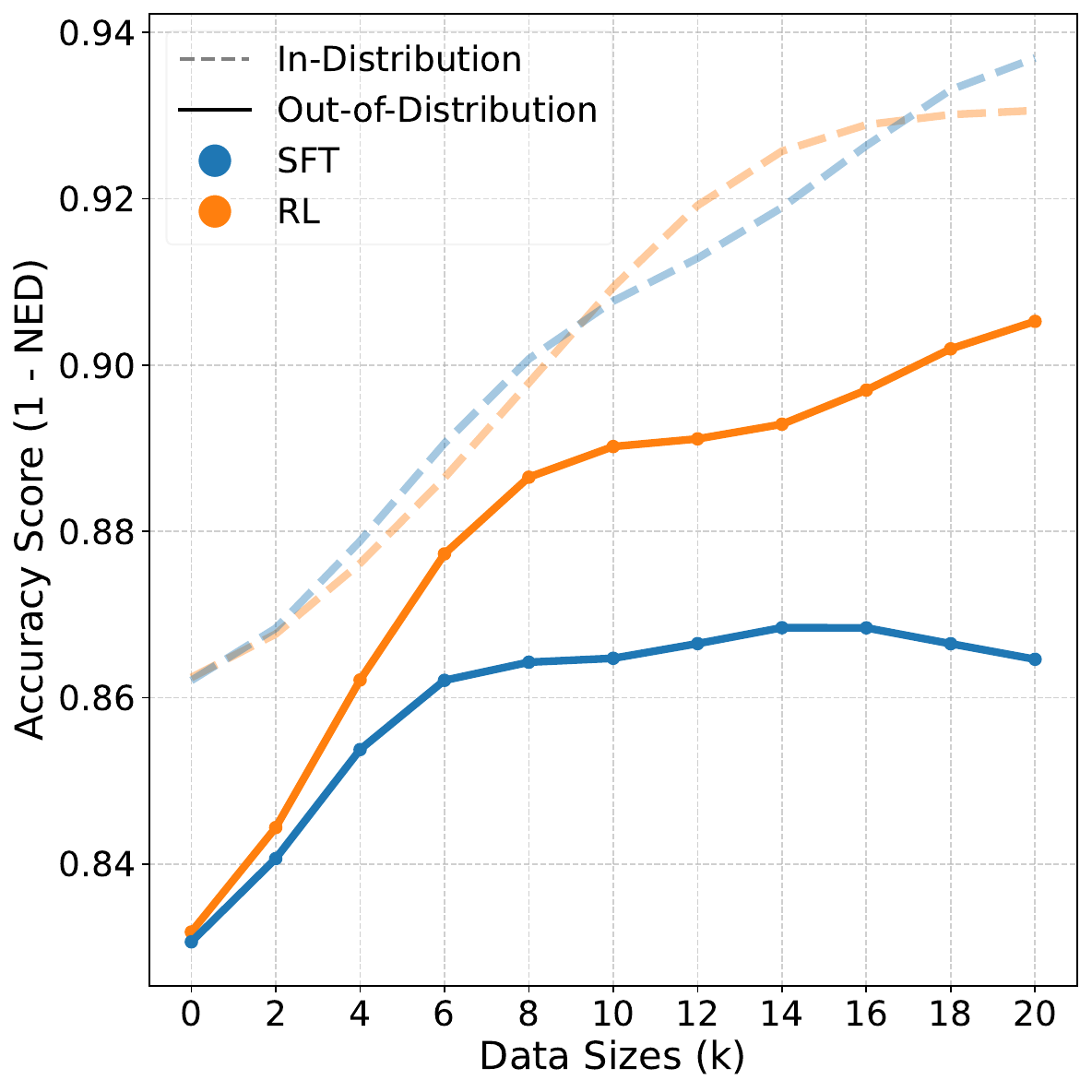}
  \vspace{-2mm}
 \caption{Comparison of document parsing performance on OmniDocBench under different training strategies as training data size increases. Left:  Evaluation with two complementary metrics: (1) Paragraph-level accuracy (edit distance evaluation on element contents only), which assesses element-wise consistency within individual element contents, independent of inter-element reading order; and (2) Page-level accuracy (edit distance evaluation on element contents and reading order), which measures global document reconstruction quality by aligning predicted outputs (e.g., texts, tables, and formulas) with ground-truth sequences. Right: In-Distribution and Out-of-Distribution task performance measured by accuracy score (1 -- NED). See detailed descriptions of the task in Section~\ref{subsec:further_analysis}.}
  \label{fig:example1223xx}
  \vspace{-5mm}
\end{figure}

% To address these limitations, we propose LayoutRL, the first end-to-end reinforcement learning framework that makes document parsers explicitly layout-aware, and construct a new dataset, as shown in Figure 1. 

To address the challenge of effectively applying RL algorithms to document parsing tasks and the lack of large-scale, high-quality data in industry to support this training process, we propose LayoutRL, the first end-to-end reinforcement learning framework for layout-aware document parsing, and construct Infinity-Doc-400K, a large-scale dataset. Specifically, our approach does not rely on explicit reasoning processes, but rather treats the entire document parsing result as the final answer and guides model learning through carefully designed reward mechanisms. We introduce verifiable rewards~\citep{shen2025vlm,liu2024deepseek,shao2024deepseekmath,zheng2025easyr1}, which consist of Edit Distance Reward~\citep{levenshtein1966binary} and Layout Parsing Reward signals, enforcing fine-grained alignment between predictions and ground-truth layouts and providing document-level supervision beyond token-level signals to encourage the model to learn transferable structural representations. Additionally, we construct a 400K-document corpus providing large-scale, high-quality supervision. It combines (1) high-fidelity synthetic scanned document parsing data—generated via HTML templates and browser rendering—and (2) expert-filtered real-world samples, pseudo-labeled through a cross-model agreement pipeline to capture genuine layout diversity. Built on this dataset, we train Infinity-Parser, an end-to-end VLM-based parser that directly outputs structured document representations.

To demonstrate the effectiveness of our approach, we conduct extensive experiments across multiple document parsing benchmarks. Results in Figure~\ref{fig:example1223xx} show that while SFT achieves strong paragraph-level accuracy, its page-level performance struggles to improve as data size increases, reflecting its tendency to memorize surface patterns rather than model the hierarchical dependencies between document structures and elements. In contrast, our layout-aware reinforcement learning method significantly outperforms SFT in page-level accuracy, while maintaining competitive paragraph-level performance. % and exhibiting smoother, more stable learning curves. 
These findings demonstrate that verifiable, layout-aware rewards enable models to move beyond simple token imitation, achieving consistent improvements in both local detail fidelity and global structural understanding, thereby establishing a more robust paradigm for document parsing. Moreover, our method also exhibits strong generalization to unseen task types in downstream applications. Finally, our method, Infinity-Parser, achieves new state-of-the-art results on four diverse benchmarks—OmniDocBench, olmOCR, PubTabNet, and FinTabNet—highlighting its effectiveness and strong cross-domain generalization.

We make the following contributions:

\begin{itemize}[left=0pt]

    \item We propose LayoutRL, a new reinforcement learning framework for end-to-end scanned document parsing, which explicitly trains models to be layout-aware by optimizing verifiable, multi-aspect rewards. Our multi-aspect reward design combines normalized edit distance, paragraph count accuracy, and reading order preservation, improving structural robustness.
  
    \item We introduce Infinity-Doc-400K, a large-scale dataset of 400,482 scanned documents that combines high-quality synthetic data with diverse real-world samples. The dataset features rich layout variations and comprehensive structural annotations, enabling robust training.
    
    \item We train a VLM based model, Infinity-Parser, which sets new state-of-the-art performance across English and Chinese benchmarks for OCR, table and formula extraction, and reading-order detection—demonstrating substantial gains in both structural fidelity and semantic accuracy over specialist pipelines and general-purpose vision-language models. 
\end{itemize}

\section{Related Work}

\subsection{Reinforcement Learning for Language Models}

Recent advancements in Large Language Models (LLMs) such as OpenAI's GPT series~\citep{openai2024gpt4ocard}), DeepSeek-R1~\citep{guo2025deepseek}, and Gemini~\citep{gemini} have highlighted the significant potential of Reinforcement Learning (RL) in enhancing their reasoning capabilities. This RL paradigm has been successfully extended to other domains demanding sophisticated reasoning, including code generation~\citep{alphacode, zeng2025acecoder}, autonomous tool utilization~\citep{toolformer, autocode}, and information retrieval~\citep{webgpt}. Similarly, RL has demonstrated its efficacy in the domain of Visual Language Models (VLMs), including precise object counting~\citep{peng2025skywork}, nuanced visual perception~\citep{liu2025visual}, and complex multimodal reasoning (e.g., VL-Rethinker~\citep{vlrethinker}, Pixel Reasoner~\citep{su2025pixel}, Vision-R1~\citep{huang2025visionr1}). These pioneering works have predominantly relied on binary outcome rewards to guide RL training. Complementary to these efforts, our work demonstrates the effectiveness of incorporating layout-aware and layout-based rewards for document parsing, offering a more granular and contextually relevant feedback mechanism. 

% The concept of curriculum learning~\citep{bengio2009curriculum, wu2024curriculum, curriculum}  also witnessed a resurgence for enhancing training efficiency and effectiveness in LLMs~\citep{kimi-1_5}. Building upon this principle, we extend the idea of curriculum learning to document parsing. Specifically, we propose leveraging perplexity as a metric for defining curriculums, thereby guiding the model to learn from progressively more challenging examples. 

\subsection{VLM-based Document Parsing}

Recent advancements in document understanding and optical character recognition (OCR) have highlighted their importance as critical benchmarks for evaluating the perceptual capabilities of vision-language models (VLMs). By incorporating large-scale OCR corpora during pretraining, models such as GPT-4o~\citep{GPT4} and Qwen2-VL~\citep{Qwen-VL} have achieved competitive performance on document content extraction tasks. Building upon these foundations, the emergence of  VLMs has further accelerated the progress of end-to-end document parsing, giving rise to a range of models such as Donut~\citep{Nougat}, Nougat~\citep{blecher2023nougat}, Kosmos-2.5~\citep{KOSMOS-2.5}, Vary~\citep{Vary}, mPLUG-DocOwl~\citep{hu2024mplug2}, Fox~\citep{fox}, and GOT~\citep{got2}. These models have continued to improve their understanding of visual layouts and textual content by leveraging advancements in visual encoders~\citep{dosovitskiy2020image}, language decoders~\citep{Qwen-VL}, and data construction pipelines. Despite the success of these VLM-based approaches in enabling end-to-end document parsing, they still face generalization challenges on downstream layout parsing tasks~\citep{MinerU}. To address this issue, we propose leveraging reinforcement learning to provide a more effective training paradigm that better aligns with the demands of document parsing.

\begin{table}[t!]
% \centering
\resizebox{1\textwidth}{!}{
\begin{tabular}{l|c|ccccc|cccc|c}
\toprule
\textbf{\multirow{2}{*}{Benchmark}} & \multirow{2}{*}{\makecell{\textbf{Document}\\\textbf{Domain}}} & \multicolumn{5}{c}{\textbf{Annotation Type}} & \multicolumn{4}{c}{\textbf{End-to-End Task}} & \multirow{2}{*}{\textbf{Exactly Match}} \\
& & BBox & Text & Table & Formula & Attributes & OCR & TR & MFR & ROD & \\

\midrule
\multicolumn{12}{l}{\textbf{\textit{End-to-end Eval Benchmarks}}} \\
\midrule
Fox~\citep{fox} & 2 & \CheckmarkBold & \CheckmarkBold &  &  & & \CheckmarkBold &  &  &  & \\
Nougat~\citep{Nougat} & 1 &  & \CheckmarkBold & \CheckmarkBold & \CheckmarkBold &  & \CheckmarkBold & \CheckmarkBold & \CheckmarkBold  &  & \\
GOT OCR 2.0~\citep{got2} & 2 &  & \CheckmarkBold & \CheckmarkBold & \CheckmarkBold &  & \CheckmarkBold & \CheckmarkBold & \CheckmarkBold  &  & \CheckmarkBold  \\
% READoc~\citep{READoc} & 2 &  & \CheckmarkBold & \CheckmarkBold & \CheckmarkBold & \CheckmarkBold & \CheckmarkBold & \CheckmarkBold & \CheckmarkBold & \CheckmarkBold & \CheckmarkBold \\

OmniDocBench~\citep{ouyang2024omnidocbench} & 9 & \CheckmarkBold & \CheckmarkBold & \CheckmarkBold & \CheckmarkBold & \CheckmarkBold & \CheckmarkBold & \CheckmarkBold & \CheckmarkBold & \CheckmarkBold & \CheckmarkBold \\

\midrule
\multicolumn{12}{l}{\textbf{\textit{End-to-end Train Dataset}}} \\
\midrule

DocStruct4M~\citep{hu2024mplug} & - &  & \CheckmarkBold &   &   &  & \CheckmarkBold &   &   &   &  \\
% MonkeyOCR~\citep{ } & - &  & \CheckmarkBold & \CheckmarkBold  & \CheckmarkBold  &  & \CheckmarkBold & \CheckmarkBold  & \CheckmarkBold  &   &  \\

olmOCR-mix~\citep{poznanski2025olmocr} & - &  & \CheckmarkBold & \CheckmarkBold & \CheckmarkBold &        
                                                                      
                                          & \CheckmarkBold & \CheckmarkBold & \CheckmarkBold & \CheckmarkBold &  \\

% \midrule
\textbf{Infinity-Doc-400K} & 7 & \CheckmarkBold & \CheckmarkBold & \CheckmarkBold & \CheckmarkBold & \CheckmarkBold & \CheckmarkBold & \CheckmarkBold & \CheckmarkBold & \CheckmarkBold & \CheckmarkBold \\

\bottomrule
\end{tabular}
}
% \vspace{-1mm}
\caption{A comparison between Infinity-Doc-400K and existing datasets. \textit{BBox}: Bounding boxes. \textit{Text}: Text in Unicode. \textit{Table}: Table in LaTeX/HTML/Markdown. \textit{Formula}: Formula in LaTeX. \textit{Attributes}: Page- and BBox-Level Attributes. \textit{OCR}: Optical Character Recognition; \textit{TR}: Table Recognition; \textit{MFR}: Math Formula Recognition; \textit{ROD}: Reading Order Detection. \textit{Multi-Type Doc}: Whether the dataset includes documents from multiple domains or categories.}
\label{tab:all-1}
\vspace{-3mm}
\end{table}

\section{Methodology}
In this section, we first introduce Infinity-Doc-400K, our large-scale multimodal dataset for end-to-end scanned document parsing. We then describe our rule-based multi-aspect reward framework, which integrates edit distance, paragraph count, and order criteria under a unified reinforcement learning objective optimized via Group Relative Policy Optimization (GRPO). 

% Finally, we detail a curriculum learning strategy that schedules training samples by structural difficulty to stabilize and accelerate convergence on complex documents.

\subsection{Infinity-Doc-400K and Generation Pipelines }

We introduce Infinity-Doc-400K, a large-scale, multimodal dataset of 400,066 richly annotated documents for end-to-end scanned document parsing. Unlike prior benchmarks that target isolated subtasks (e.g., layout detection, OCR, or table recognition), Infinity-Doc-400K provides holistic supervision by pairing rendered scanned document pages with their ground-truth Markdown representations. This design enables training and evaluating models that directly translate visual inputs to layout outputs without relying on brittle, multi-stage pipelines. As shown in Table \ref{tab:all-1}, compared to existing works, Infinity-Doc-400K not only significantly enhances task diversity but also substantially improves overall data quality through our proposed synthetic generation mechanism. More details on the data distribution and quality control are provided in the Data Details section of the Appendix.

\begin{figure}[t]
  \centering
  % 插入图片，options 中可以设置宽度、高度、缩放比例等
  \includegraphics[width=1\linewidth]{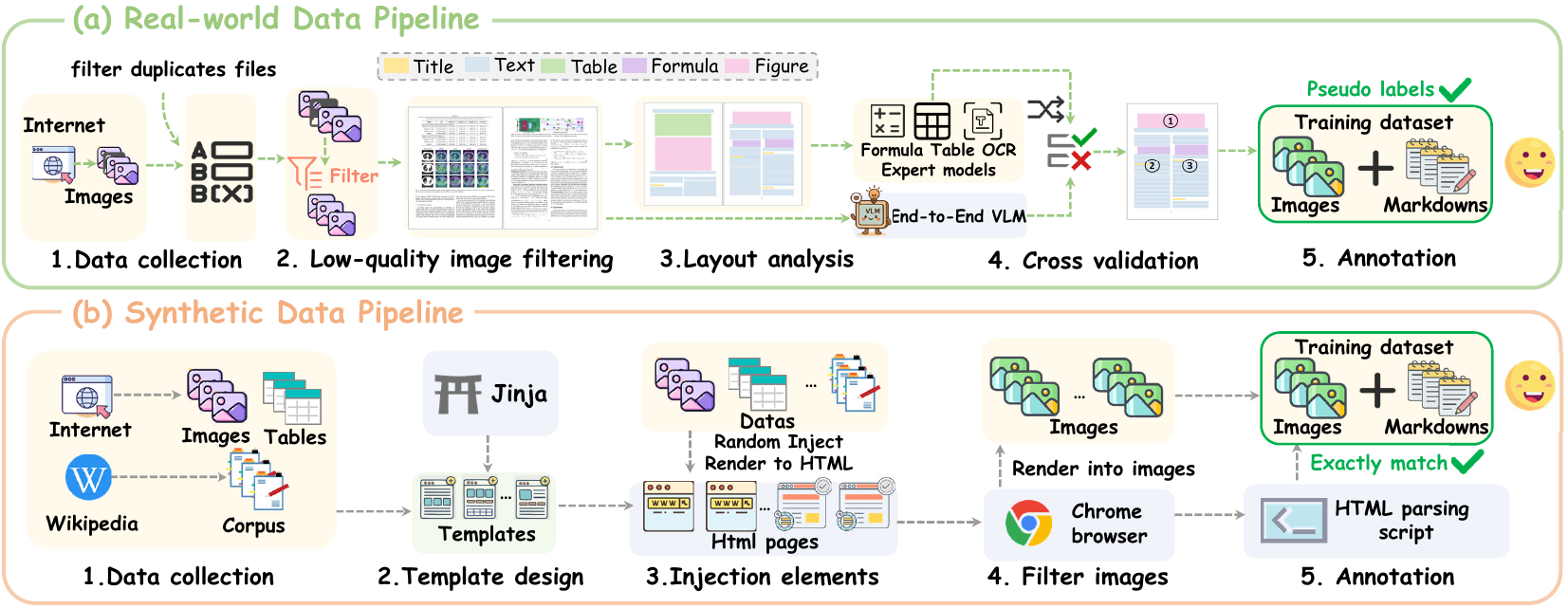}
  \vspace{-2mm}
  \caption{ Data construction pipelines for document parsing.  (a) Real-world pipelines enhance quality by combining multiple expert models and layout analysis, yielding better-aligned supervision through intersection and reading order reasoning. (b) Synthetic pipeline leverages structured HTML templates and browser rendering to generate clean, exactly-aligned scanned document parsing data, ensuring high-quality supervision for end-to-end parsing. }
  \label{fig:example111}
% \vspace{-3mm}
\end{figure}

To construct Infinity-Doc-400K, we design a dual-pipeline framework that integrates both synthetic and real-world document generation, as illustrated in Figure~\ref{fig:example111}. This design addresses a critical limitation of traditional data construction pipelines, which often rely on weak supervision and pseudo-labeling from a single model applied to crawled, scanned documents. These pipelines frequently suffer from noisy, misaligned, or incomplete annotations, especially in complex layouts or multilingual content, thus hindering model performance and generalization. To overcome these issues, our dual-pipeline framework is motivated by the need to balance annotation quality and structural diversity. The synthetic branch provides highly accurate, clean, and precisely aligned annotations at scale, while the real-world branch introduces naturally occurring layout variability and semantic richness, which are essential for building models that generalize robustly in practical applications.

\paragraph{Real-World Data} We develop a real-world data construction pipeline to capture the structural complexity and natural layout variability of documents across practical domains. We collect diverse scanned documents from sources such as financial reports, medical records, academic papers, books, magazines, and web pages, covering both dense and sparse content layouts. To generate structural annotations, we adopt a multi-expert strategy, where specialized models handle different structural elements, such as layout blocks, texts, formulas, and tables. For example, overall layouts are analyzed by a visual layout model~\citep{huang2022layoutlmv3pretrainingdocumentai}, formula regions are processed by a dedicated formula recognition model~\citep{wang2024unimernetuniversalnetworkrealworld}, and tables are parsed by a transformer-based table extractor~\citep{Nougat}. A cross-validation mechanism is then applied to filter out inconsistencies by comparing the outputs of expert models and VLMs. Only regions with consistent predictions across models are retained as high-confidence pseudo-ground-truth annotations. This layout-aware filtering results in a rich and reliable dataset that reflects the complexity of real-world documents and supports robust document parsing model training.

\paragraph{Synthetic Data} We design a synthetic data construction pipeline. We collect text and images from sources such as Wikipedia, web crawlers, and online corpora, and use Jinja~\citep{nipkow2003jinja} templates to inject sampled content into predefined single-, double-, or triple-column HTML layouts. These pages are rendered into scanned documents using a browser engine, followed by automated filtering to remove low-quality or overlapping images. Ground-truth annotations are extracted by parsing the original HTML to produce aligned Markdown representations.  This synthetic approach not only significantly reduces construction costs and ensures annotation accuracy and structural diversity, but more importantly, it addresses the longstanding issue of imprecise or inconsistent supervision commonly found in pseudo-labeled datasets, providing high-quality and well-aligned supervision for training end-to-end models.

\paragraph{Quality Control Measures} To ensure reliable annotations at scale, we designed a hybrid quality control strategy for Infinity-Doc-400K. First, three domain experts with doctoral degrees in document analysis manually inspected about 5\% of the data. Their feedback not only identified potential errors but also served as a quality anchor for evaluating annotation consistency. Guided by these inspections, we iteratively refined the screening rules no fewer than five times, continuously improving the labeling pipeline. Finally, to achieve scalability, we employed a model-based cross-verification mechanism: multiple models generated annotations for real-world samples, with high-consistency outputs retained and inconsistent cases fed back into further rule refinement. This layered framework—anchored by expert inspection, strengthened through iterative rule optimization, and scaled via model cross-checking—effectively balances annotation reliability with dataset scalability.

\subsection{RL with Layout-Aware Rewards}
\begin{figure}[t]
  \centering
  % 插入图片，options 中可以设置宽度、高度、缩放比例等
  \includegraphics[width=0.98\linewidth]{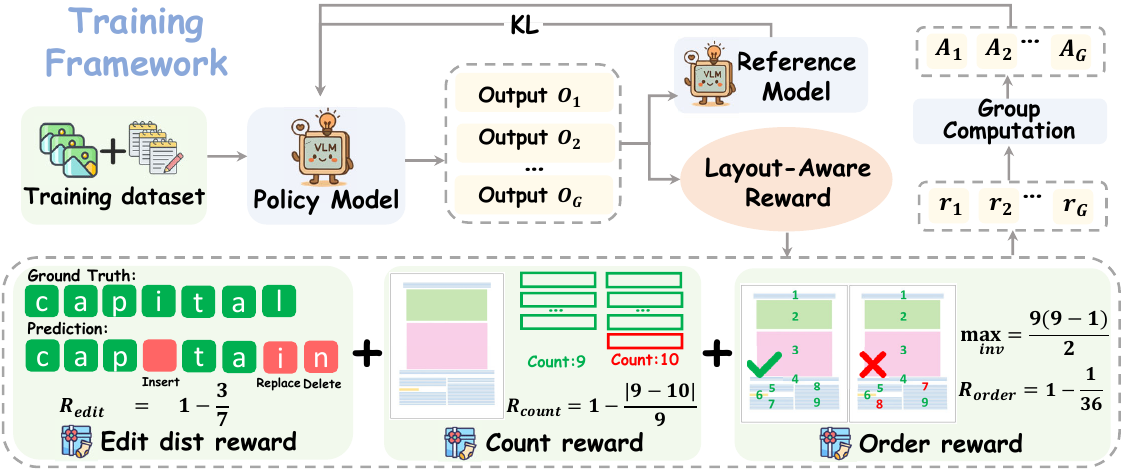}
  % \vspace{-1mm}
  \caption{Overview of Infinity-Parser training framework. Our model is optimized via reinforcement finetuning with edit distance, layout, and order-based rewards. 
  }
  \label{fig:example222}
  % \vspace{-5mm}
\end{figure}

As illustrated in Figure~\ref{fig:example222}, we employ an RL framework to directly optimize scanned document parsers, aiming to enhance both structural fidelity and semantic accuracy. Specifically, we utilize GRPO~\citep{shao2024deepseekmath}, which enables learning from rule-based reward signals without relying on absolute values. GRPO operates by generating a set of candidate Markdown outputs for each document and evaluating them using a \textit{multi-aspect reward}, denoted as $R_{\text{Multi-Aspect}}$, which integrates multiple rule-based criteria into a unified supervisory signal. These raw rewards are then converted into relative advantage scores by comparing each candidate against others within the same group. This relative evaluation promotes training stability and encourages the selection of higher-quality outputs, eliminating the need for a learned value function or critic. Notably, our reinforcement learning approach avoids any explicit thinking or intermediate reasoning process; instead, all outputs are treated as final answers, with the model receiving verifiable rewards based on these outputs.

The multi-aspect reward $R_{\text{Multi-Aspect}}$ consists of three complementary components, each capturing a different aspect of parsing quality:

\paragraph{Edit Distance Reward ($R_{\mathrm{dist}}$)}  
We define the edit distance reward based on the normalized Levenshtein distance \( D(y, \hat{y}) \) between the predicted output \( \hat{y} \) and the reference output \( y \):

\vspace{-3mm}
\begin{equation}
R_{\text{dist}} = 1 - \tfrac{D(y, \hat{y})}{\max(N, M)}
\end{equation}
\vspace{-3mm}

where \( N = |y| \) and \( M = |\hat{y}| \) are the lengths of the reference and predicted sequences, respectively. The distance \( D(y, \hat{y}) \) measures the minimum number of single-character insertions, deletions, or substitutions required to convert \( \hat{y} \) into \( y \), thereby capturing both semantic and formatting discrepancies. This reward is bounded within \([0, 1]\), with higher values indicating better alignment between prediction and reference. Meanwhile, the reference output \( y \) is synthesized through two data generation pipelines proposed in this work. It is produced with rigorous rule-based filtering and consistency validation, and serves as a high-quality surrogate for ground-truth annotations in evaluating the quality of the model output.

\paragraph{Count Reward ($R_{\mathrm{count}}$)}
To encourage accurate paragraph segmentation, let $N_Y$ and $N_{\hat Y}$ be the numbers of reference and predicted paragraphs. We define:

\vspace{-3mm}
\begin{equation}
R_{\text{count}} = 1 - \tfrac{|N_Y - N_{\hat{Y}}|}{N_Y}
\end{equation}
\vspace{-3mm}
 
which penalizes missing or spurious paragraphs.

\paragraph{Order Reward ($R_{\mathrm{order}}$)}
We measure sequence-level fidelity by counting pairwise inversions $D_{\mathrm{order}}$ between reference and predicted paragraphs. with $\max_{\mathrm{inv}}=N_Y(N_Y-1)/2$, we set:

\vspace{-3mm}
\begin{equation}
R_{\text{order}} = 1 - \frac{D_{\text{order}}}{\max_{\text{inv}}}
\end{equation}

rewarding preservation of the original reading order.

The final multi-aspect reward is a weighted combination of these three components. Specifically, we begin by applying the Hungarian algorithm~\citep{kuhn1955hungarian} to establish the optimal one-to-one matching between predicted and ground-truth segments, identifying both pairings and their relative order. Based on the matched segment count, we compute the count reward to reflect alignment in the number of segments. Using the relative sequence of matched pairs, we calculate the order reward to measure structural consistency. On top of these matchings, we compute the edit reward by averaging the edit similarities of each matched segment pair.  Combining these terms yields the final reward:

\vspace{-3mm}
\begin{equation}
R_{\text{Multi-Aspect}} = R_{\text{dist}} + R_{\text{count}} + R_{\text{order}}
\end{equation}

This multi-aspect design balances content fidelity with structural correctness and order preservation, providing rich supervision for end-to-end document parsing.

\section{Experiments}

% \subsection{Implementation Details}
% We adopt Qwen2.5-VL-7B~\citep{bai2025qwen2} as the base model and apply the VeRL~\citep{sheng2024hybridflow} framework for reinforcement learning. Detailed implementation can be found in the Appendix.

\subsection{Implementation Details}

% We fine-tune Qwen2.5-VL-7B using GRPO under a distributed training setup based on DeepSpeed ZeRO-3. The model is trained on 8 A100 GPUs (80GB) using %DeepSpeed stage 3 with offloading disabled, and activation checkpointing enabled to reduce memory footprint. We use the AdamW optimizer with $(\beta_1 = %0.9, \beta_2 = 0.95)$ and weight decay set to 0.1. The learning rate is initialized at 5e-6 with a linear warmup of 500 steps followed by cosine decay. The %batch size per GPU is 1, resulting in an effective batch size of 8, and the maximum sequence length is set to 8192 tokens. Our final experimental results %were obtained after training for 3 days on 8 NVIDIA A100 GPUs.

We fine-tune the Qwen2.5-VL-7B model using GRPO within a distributed training setup based on Verl~\cite{sheng2024hybridflow, zheng2025easyr1}, utilizing 8 A100 GPUs (80GB). Throughout our experiments, we set the KL coefficient $ \beta= 1.0 \times 10^{-2} $. And for each problem instance, we sample 8 responses, each with a maximum length of 8192 tokens and a temperature of 1.0. Both the rollout batch size and the global batch size are set to 128. The actor model is updated using the AdamW optimizer with parameters $(\beta_1 = 0.9, \beta_2 = 0.99)$ and a learning rate $1.0 \times 10^{-6}$. The model is trained for 1.0 epoch for all experiments. Due to limited computational resources, we randomly sampled 43K documents from the 400K corpus for training. In our main results, we directly performed reinforcement learning on the base model using the 43K subset.

\begin{table*}[t]
  \begin{threeparttable}
    \resizebox{1\textwidth}{!}{%
      \begin{tabular}{l cc | cc  cc cc cc cc }
      \toprule
      \textbf{Methods}
        & \multicolumn{2}{c}{\textbf{Overall\textsuperscript{Edit}}$\downarrow$} 
        & \multicolumn{2}{c}{\textbf{Text\textsuperscript{Edit}}$\downarrow$} 
        & \multicolumn{2}{c}{\textbf{Form.\textsuperscript{Edit}}$\downarrow$} 
        & \multicolumn{2}{c}{\textbf{Table\textsuperscript{TEDS}}$\uparrow$} 
        & \multicolumn{2}{c}{\textbf{Table\textsuperscript{Edit}}$\downarrow$}
        & \multicolumn{2}{c}{\textbf{Read Order\textsuperscript{Edit}}$\downarrow$} \\
      \cmidrule{2-13}
       & {\it EN} & {\it ZH} 
       & {\it EN} & {\it ZH} 
       & {\it EN} & {\it ZH} 
       & {\it EN} & {\it ZH} 
       & {\it EN} & {\it ZH} 
       & {\it EN} & {\it ZH} \\
      \midrule
      \multicolumn{6}{l}{\textbf{Based on Pipeline Tools}} \\
    
      MinerU~\citep{MinerU} 
        & 0.15 & 0.357
        & \textbf{0.061} & 0.215
        & \textbf{0.278} & 0.577
        & 78.6 & 62.1
        &  0.18 & 0.344
        & 0.079 & 0.292 \\
      Marker~\citep{marker}
        & 0.336 & 0.556
        & 0.080 & 0.315
        & 0.530 & 0.883
        & 67.6 & 49.2
        & 0.619 & 0.685
        & 0.114 & 0.340 \\
      Mathpix
        & 0.191 & 0.365
        & 0.105 & 0.384
        & 0.306 & 0.454
        & 77.0 & 67.1
        & 0.243 & 0.320
        & 0.108 & 0.304 \\
    
    Docling~\citep{livathinos2025docling}
        & 0.589 & 0.909
        & 0.416 & 0.987
        & 0.999 & 1.000
        & 61.3 & 25.0
        & 0.627 & 0.810
        & 0.313 & 0.837 \\
    Pix2Text~\citep{gurgurov2024image}
        & 0.320 & 0.528
        & 0.138 & 0.356
        & 0.276 & 0.611
        & 73.6 & 66.2
        & 0.584 & 0.645
        & 0.281 & 0.499 \\
    Unstructured-0.17.2
        & 0.586 & 0.716
        & 0.198 & 0.481
        & 0.999 & 1.000
        & - & -
        & 1.000 & 0.998
        & 0.145 & 0.387 \\
    OpenParse-0.7.0
        & 0.646 & 0.814
        & 0.681 & 0.974
        & 0.996 & 1.000
        & 64.8 & 27.5
        & 0.284 & 0.639
        & 0.595 & 0.641 \\
    \midrule
    \multicolumn{6}{l}{\textbf{Based on Expert VLMs}} \\
      GOT-OCR~\citep{got2} 
        & 0.287 & 0.411
        & 0.189 & 0.315 
        & 0.360 & 0.528
        & 53.2 & 47.2
        & 0.459 & 0.520
        & 0.141 & 0.280 \\
      Nougat~\citep{Nougat} 
        & 0.452 & 0.973
        & 0.365 & 0.998 
        & 0.488 & 0.941
        & 39.9 & 0.0
        & 0.572 & 1.000
        & 0.382 & 0.954 \\
    Mistral OCR
        & 0.268 & 0.439
        & 0.072 & 0.325
        & 0.318 & 0.495
        & 75.8 & 63.6
        & 0.600 & 0.650
        & 0.083 & 0.284 \\
    OLMOCR-sglang
        & 0.326 & 0.469
        & 0.097 & 0.293
        & 0.455 & 0.655
        & 68.1 & 61.3
        & 0.608 & 0.652
        & 0.145 & 0.277 \\
    SmolDocling-256M
        & 0.493 & 0.816
        & 0.262 & 0.838
        & 0.753 & 0.997
        & 44.9 & 16.5
        & 0.729 & 0.907
        & 0.227 & 0.522 \\
        
    \midrule
    \multicolumn{6}{l}{\textbf{Based on General VLMs}} \\
      GPT-4o~\citep{GPT4} 
        & 0.233 & 0.399
        & 0.144 & 0.409
        & 0.425 & 0.606
        & 72.0 & 62.9
        & 0.234 & 0.329
        & 0.128 & 0.251 \\
    
    Qwen2-VL-72B~\citep{wang2024qwen2}   
        & 0.252 & 0.327
        & 0.096 & 0.218
        & 0.404 & 0.487
        & 76.8 & 76.4
        & 0.387 & 0.408
        & 0.119 & 0.193 \\
    InternVL2-76B~\citep{InternVL}  
        & 0.440 & 0.443
        & 0.353 & 0.290
        & 0.543 & 0.701
        & 63.0 & 60.2
        & 0.547 & 0.555
        & 0.317 & 0.228 \\ 

    Qwen2.5-VL-7B~\citep{bai2025qwen2.5}
        & 0.220 & 0.265
        & 0.142 & 0.205
        & 0.393 & 0.530
        & 78.7 & 78.3
        & 0.155 & 0.162
        & 0.191 & 0.169 \\
    InternVL3-8B~\citep{zhu2025internvl3}
        & 0.426 & 0.385
        & 0.315 & 0.345
        & 0.714 & 0.729
        & 59.0 & 71.5
        & 0.352 & 0.211
        & 0.324 & 0.257 \\
    \midrule
    \multicolumn{6}{l}{\textbf{Based on Reinforcement Learning }} \\
    Infinity-Parser-7B
        & \textbf{0.141} & \textbf{0.197} 
        & 0.076  & \textbf{0.117 }
        & 0.314  & \textbf{0.434 }
        & \textbf{85.3} & \textbf{81.4}
        & \textbf{0.098} & \textbf{0.142}
        & \textbf{0.076} & \textbf{0.095} \\
      \bottomrule
      \end{tabular}%
    }% end resizebox
  \end{threeparttable}
  \caption{Comprehensive evaluation of document parsing algorithms on OmniDocBench: performance metrics for text, formula, table, and reading order extraction, with overall scores derived from ground truth comparisons. }
  \label{result_label1}
  % \vspace{-3mm}
\end{table*}

\subsection{Main Results}

% \subsection{Evaluation and Metrics}
% In this section, we evaluate the model's performance in document parsing, table recognition, and Document-level OCR. In this work, 

We evaluate our method on several widely-used benchmarks for document understanding and OCR tasks. OmniDocBench~\citep{ouyang2024omnidocbench} provides comprehensive evaluation across diverse document types using NED and TEDS metrics. % Fox~\citep{liu2024focus_fox} offers multilingual assessment with 212 bilingual pages covering 9 sub-tasks. 
For table recognition, we use PubTabNet with scientific tables and FinTabNet~\citep{zheng2021global} with financial documents. Additionally, we employ olmOCR-Bench~\citep{poznanski2025olmocr} for fact-based OCR evaluation. We ensure that the test data for each benchmark undergoes rigorous text similarity filtering to prevent any overlap with the training data.  Detailed descriptions of these benchmarks are provided in Appendix.

\paragraph{Overall Evaluation on OmniDocBench}
As shown in Table \ref{result_label1}, pipeline-based methods such as MinerU~\citep{MinerU} and Mathpix achieve superior performance across individual sub-tasks including text recognition and formula recognition. Meanwhile, general-purpose vision-language models like Qwen2.5-VL-7B and GPT-4o also demonstrate competitive results. Notably, most methods perform better on English pages compared to Chinese pages, reflecting language-dependent challenges. In contrast, our proposed Infinity-Parser-7B achieves a more balanced performance across all sub-tasks and languages, setting new SOTA results with overall edit distances of 0.141 and 0.197. This highlights the advantage of reinforcement learning with multi-aspect rewards in enabling robust, end-to-end document parsing.

\begin{table}[h]
\centering
\resizebox{\linewidth}{!}{
\begin{tabular}{l c|cccccccc}
\toprule
\textbf{Model} & \textbf{Overall} & \textbf{ArXiv} & \textbf{Old Scans Math} & \textbf{Tables} & \textbf{Old Scans} & \textbf{Headers\&Footers} & \textbf{Multi Col.} & \textbf{Long-Tiny Text} & \textbf{Base} \\
\midrule
GOT OCR & 48.3 & 52.7 & 52.0 & 0.2 & 22.1 & 93.6 & 42.0 & 29.9 & 94.0 \\
Marker v1.6.2 & 59.4 & 24.3 & 22.1 & 69.8 & 24.3 & 87.1 & 71.0 & 76.9 & 99.5 \\
MinerU v1.3.10 & 61.5 & 75.4 & 47.4 & 60.9 & 17.3 & \textbf{96.6} & 59.0 & 39.1 & 96.6 \\
Mistral OCR API & 72.0 & 77.2 & 67.5 & 60.6 & 29.3 & 93.6 & 71.3 & 77.1 & 99.4 \\
GPT-4o (No Anchor) & 68.9 & 51.5 & 75.5 & 69.1 & 40.9 & 94.2 & 68.9 & 54.1 & 96.7 \\
GPT-4o (Anchored) & 69.9 & 53.5 & 74.5 & 70.0 & 40.7 & 93.8 & 69.3 & 60.6 & 96.8 \\
Gemini Flash 2 (No Anchor) & 57.8 & 32.1 & 56.3 & 61.4 & 27.8 & 48.0 & 58.7 & 84.4 & 94.0 \\
Gemini Flash 2 (Anchored) & 63.8 & 54.5 & 56.1 & 72.1 & 34.2 & 64.7 & 61.5 & 71.5 & 95.6 \\
Qwen 2 VL (No Anchor) & 31.5 & 19.7 & 31.7 & 24.2 & 17.1 & 88.9 & 8.3 & 6.8 & 55.5 \\
Qwen 2.5 VL (No Anchor) & 65.5 & 63.1 & 65.7 & 67.3 & 38.6 & 73.6 & 68.3 & 49.1 & 98.3 \\
olmOCR v0.1.68 (No Anchor) & 76.3 & 72.1 & 74.7 & 71.5 & 43.7 & 91.6 & 78.5 & 80.5 & 98.1 \\
olmOCR v0.1.68 (Anchored) & 77.4 & 75.6 & 75.1 & 70.2 & 44.5 & 93.4 & 79.4 & 81.7 & 99.0 \\
\midrule
Infinity-Parser-7B & \textbf{82.5} & \textbf{84.4} & \textbf{83.8} & \textbf{85.0} & \textbf{47.9} & 88.7 & \textbf{84.2} & \textbf{86.4} & \textbf{99.8} \\
\bottomrule
\end{tabular}
}
\caption{Performance comparison on the olmOCR~\citep{poznanski2025olmocr} benchmark across multiple document domains and structural challenges. Higher is better.}
\label{tab:olmocr_results}
% \vspace{-5mm}
\end{table}

% \subsection{Document-Level OCR Evaluation on olmOCR-Bench}

\textbf{Document-level OCR Evaluation} Table \ref{tab:olmocr_results} reports performance on the olmOCR-Bench benchmark, which evaluates document-level OCR across diverse layouts and domains. Infinity-Parser-7B achieves the highest overall score (82.5), followed closely by olmOCR v0.1.68 (Anchored) (77.4), both demonstrating strong performance in complex categories like multi-column layouts and scanned math content. The results highlight the effectiveness of anchored prompting, with anchored versions of models (e.g., GPT-4o, olmOCR) significantly outperforming their non-anchored counterparts—especially on tables and old scans. This underscores the importance of layout-aware extraction techniques. In contrast, traditional pipelines like Marker and GOT OCR lag behind in structural accuracy, reinforcing the value of modern VLM-based approaches in high-fidelity PDF understanding.

\begin{table*}[h]
  % \centering
  \begin{minipage}[t]{0.47\textwidth}
    \vspace{0pt} % 对齐顶部
    % \small  
     \paragraph{Table Recognition Evaluation} To evaluate the model’s generalization ability, we introduce task-specific test cases. In Table~\ref{tab:teds_grouped_comparison}, we compare Infinity-Parser-7B with end-to-end table recognition models on PubTabNet and FinTabNet using the TEDS metric, which evaluates both structure and content. We also report TEDS-S for structure-only assessment. The evaluation results for InternVL3, Qwen2.5-VL, and GPT-4o were generated through our standardized benchmarking pipeline. Infinity-Parser-7B achieves the highest TEDS-S and TEDS scores on both datasets.
     % surpassing all baselines including strong models like InternVL3-78B and Qwen2.5-VL-72B.
  \end{minipage}
  \hfill
  \begin{minipage}[t]{0.5\textwidth}
    \vspace{0pt} % 对齐顶部
    \centering
    \small
    \renewcommand{\arraystretch}{1.1}
    \resizebox{\linewidth}{!}{%
    \begin{tabular}{lcccc}
      \toprule
      \textbf{Model} & \multicolumn{2}{c}{PubTabNet} & \multicolumn{2}{c}{FinTabNet} \\
      \cmidrule(lr){2-3} \cmidrule(lr){4-5}
      & TEDS-S & TEDS & TEDS-S & TEDS \\
      \midrule
      % EDD~\citep{zhong2020image} & 89.9 & 88.3 & 90.6 & - \\
      % OmniParser~\citep{wan2024omniparser} & 90.45 & 88.83 & 91.55 & 89.75 \\
      % InternVL3-8B~\citep{zhu2025internvl3} & 87.48 & 83.02 & 86.73 & 84.01 \\
      % InternVL3-78B~\citep{zhu2025internvl3} & 89.63 & 82.11 & 92.51 & 89.21 \\
      % Qwen2.5-VL-7B~\citep{bai2025qwen2} & 86.78 & 81.60 & 87.46 & 82.58 \\
      % Qwen2.5-VL-72B~\citep{bai2025qwen2} & 87.91 & 84.39 & 87.13 & 82.90 \\
      % GPT-4o~\citep{GPT4o} & 86.16 & 76.53 & 87.00 & 83.96 \\

      EDD & 89.9 & 88.3 & 90.6 & - \\
    OmniParser & 90.45 & 88.83 & 91.55 & 89.75 \\
    InternVL3-8B & 87.48 & 83.02 & 86.73 & 84.01 \\
    InternVL3-78B & 89.63 & 82.11 & 92.51 & 89.21 \\
    Qwen2.5-VL-7B & 86.78 & 81.60 & 87.46 & 82.58 \\
    Qwen2.5-VL-72B & 87.91 & 84.39 & 87.13 & 82.90 \\
    GPT-4o & 86.16 & 76.53 & 87.00 & 83.96 \\
      \midrule
      \textbf{Infinity-Parser-7B} & \textbf{93.46} & \textbf{91.82} & \textbf{97.16} & \textbf{95.92} \\
      \bottomrule
    \end{tabular}%
    }
    \captionof{table}{Comparisons of end-to-end table recognition methods on PubTabNet and FinTabNet.}
    \label{tab:teds_grouped_comparison}
  \end{minipage}
  % \vspace{-3mm}
\end{table*}

% \begin{table}[htbp]
%     \small
%     \centering
%     \renewcommand\arraystretch{1}
%     \setlength{\tabcolsep}{6pt}
%     \resizebox{1\linewidth}{!}{
%     \begin{tabular}{lccc cc|cc}
%         \toprule
%         \textbf{Method} & \textbf{Edit Dist. Reward} & \textbf{Count Reward} & \textbf{Order Reward} & \textbf{SFT} & \textbf{RL} & \textbf{Overall}\textsuperscript{Edit} $\downarrow$ & \textbf{Overall}\textsuperscript{Cat.} $\downarrow$ \\
%         \midrule
%         Zero Shot & - & - & - & -   & -    & 0.346 & 0.259 \\
%         SFT       & - & - & - & 27k & -    & 0.375 & 0.166 \\ 
%         \midrule
%         Zero + RL  & \CheckmarkBold & - & - & -   & 27k  & 0.287 & 0.156 \\
%         Zero + RL  & \CheckmarkBold & \CheckmarkBold & - & - & 27k & 0.280 & 0.141 \\
%         Zero + RL & \CheckmarkBold & \CheckmarkBold & \CheckmarkBold & - & 27k & 0.260 & 0.123 \\
%         \midrule
%         SFT + RL  & \CheckmarkBold & - & - & 300k  & 27k  & 0.345 & 0.244 \\
%         SFT + RL  & \CheckmarkBold & \CheckmarkBold & - & 300k  & 27k  & 0.345 & 0.244 \\
%         SFT + RL  & \CheckmarkBold & \CheckmarkBold & \CheckmarkBold & 300k  & 27k  & 0.345 & 0.244 \\
%         \bottomrule
%     \end{tabular}
%     }
%     \vspace{-2mm}
%     \caption{Results under different reward designs.}
%     \label{tab:test_rewards}
% \vspace{-3mm}
% \end{table}

\subsection{Ablation Study}

We perform ablation experiments to evaluate the individual contributions of our three core design choices: (1) data quality verification and (2) multi-aspect rewards. We report all ablation results using two primary evaluation metrics: Overall\textsuperscript{Edit} and Overall\textsuperscript{Cat.}. Overall\textsuperscript{Edit} represents the average edit-based overall score across English and Chinese pages, as shown in Table~\ref{result_label1}. In contrast, Overall\textsuperscript{Cat.} reflects the mean category-level performance across nine types of scanned document pages, following the same evaluation setting as Table 9 in the Appendix.  
% To save computational resources, the ablation studies are conducted on a randomly selected subset of 22K samples, rather than using the full 55K dataset as in the final results.

% \begin{table}[htbp]
%     \small
%     \centering
%     \renewcommand\arraystretch{1}
%     \setlength{\tabcolsep}{6pt}
%     \resizebox{1\linewidth}{!}{
%     \begin{tabular}{lccc cc|cc}
%         \toprule
%         \textbf{Method} & \textbf{Edit Dist. Reward} & \textbf{Count Reward} & \textbf{Order Reward} & \textbf{SFT} & \textbf{RL} & \textbf{Overall}\textsuperscript{Edit} $\downarrow$ & \textbf{Overall}\textsuperscript{Cat.} $\downarrow$ \\
%         \midrule
%         Zero Shot & - & - & - & -   & -    & 0.242 & 0.182 \\
%         SFT       & - & - & - & 43K & -    & 0.205 & 0.129 \\
%         SFT + RL  & - & - & - & 5k  & 37k  & 0.175 & 0.095 \\
%         Zero + RL  & \CheckmarkBold & - & - & -   & 43K  & 0.287 & 0.156 \\
%         Zero + RL  & \CheckmarkBold & \CheckmarkBold & - & - & 43K & 0.180 & 0.112 \\
%         Zero + RL & \CheckmarkBold & \CheckmarkBold & \CheckmarkBold & - & 43K & 0.169 & 0.106 \\
%         \bottomrule
%     \end{tabular}
%     }
%     % \vspace{-2mm}
%     \caption{Results under different reward designs.}
%     \label{tab:test_rewards}
% % \vspace{-3mm}
% \end{table}

\begin{table}[htbp]
    \small
    \centering
    \renewcommand\arraystretch{1}
    \setlength{\tabcolsep}{6pt}
    \resizebox{1\linewidth}{!}{
    \begin{tabular}{lccc cc|ccc}
        \toprule
        \textbf{Method} & \textbf{Edit Dist.} & \textbf{Count.} & \textbf{Order.} & \textbf{SFT} & \textbf{RL} & \textbf{Overall (EN)}\textsuperscript{Edit} $\downarrow$ & \textbf{Overall (ZH)}\textsuperscript{Edit} $\downarrow$ & \textbf{Overall }\textsuperscript{Cat.} $\downarrow$ \\
        \midrule
        Zero Shot & - & - & - & -   & -    & 0.220 & 0.265 & 0.183 \\
        SFT       & - & - & - & 43K & -    & 0.198 & 0.261 & 0.159 \\
        Zero + RL  & \CheckmarkBold & - & - & -   & 43K  & 0.169 & 0.224 & 0.156 \\
        Zero + RL  & \CheckmarkBold & \CheckmarkBold & - & - & 43K & 0.159 & 0.200 & 0.112 \\
        Zero + RL & \CheckmarkBold & \CheckmarkBold & \CheckmarkBold & - & 43K & 0.141 & 0.197 & 0.104 \\ 
        SFT + RL   & \CheckmarkBold & \CheckmarkBold & \CheckmarkBold & 43K & 43K    & 0.163 & 0.195 & 0.092 \\
        \bottomrule
    \end{tabular}
    } 
    \caption{Results under different reward designs.}
    \label{tab:test_rewards}
% \vspace{-5mm}
\end{table}

\paragraph{Effect of Multi-Aspect Rewards.} Table~\ref{tab:test_rewards} demonstrates that reinforcement learning can outperform supervised fine-tuning when appropriate reward designs are applied. Compared to the SFT baseline (0.198 / 0.261 / 0.159), the RL method with distance-based reward (\textit{Zero + $R_{\text{dist}}$}) achieves better Overall\textsuperscript{Edit} (EN: 0.169 vs. 0.198, ZH: 0.224 vs. 0.261) while maintaining a comparable Overall\textsuperscript{Cat.} (0.156 vs. 0.159). Incorporating additional count and order rewards further improves structural consistency: \textit{Zero + $R_{\text{dist}} + R_{\text{count}}$} achieves 0.159 / 0.200 / 0.112, and \textit{Zero + $R_{\text{dist}} + R_{\text{count}} + R_{\text{order}}$} achieves 0.141 / 0.197 / 0.104. Meanwhile, when reinforcement learning is combined with supervised fine-tuning (\textit{SFT + RL}), the model does not exhibit further significant improvements. This observation is consistent with existing studies~\citep{guo2025deepseek}, which suggest that given the backbone model’s inherent capabilities, SFT may be unnecessary for downstream RL training in certain scenarios. These results further demonstrate that reinforcement learning, when equipped with structural supervisory signals, enables the model to better align with task-specific objectives.

\begin{figure}[h!]
    \centering
    % 子图1
    \begin{subfigure}{0.49\textwidth}
        \centering
        \includegraphics[width=\linewidth]{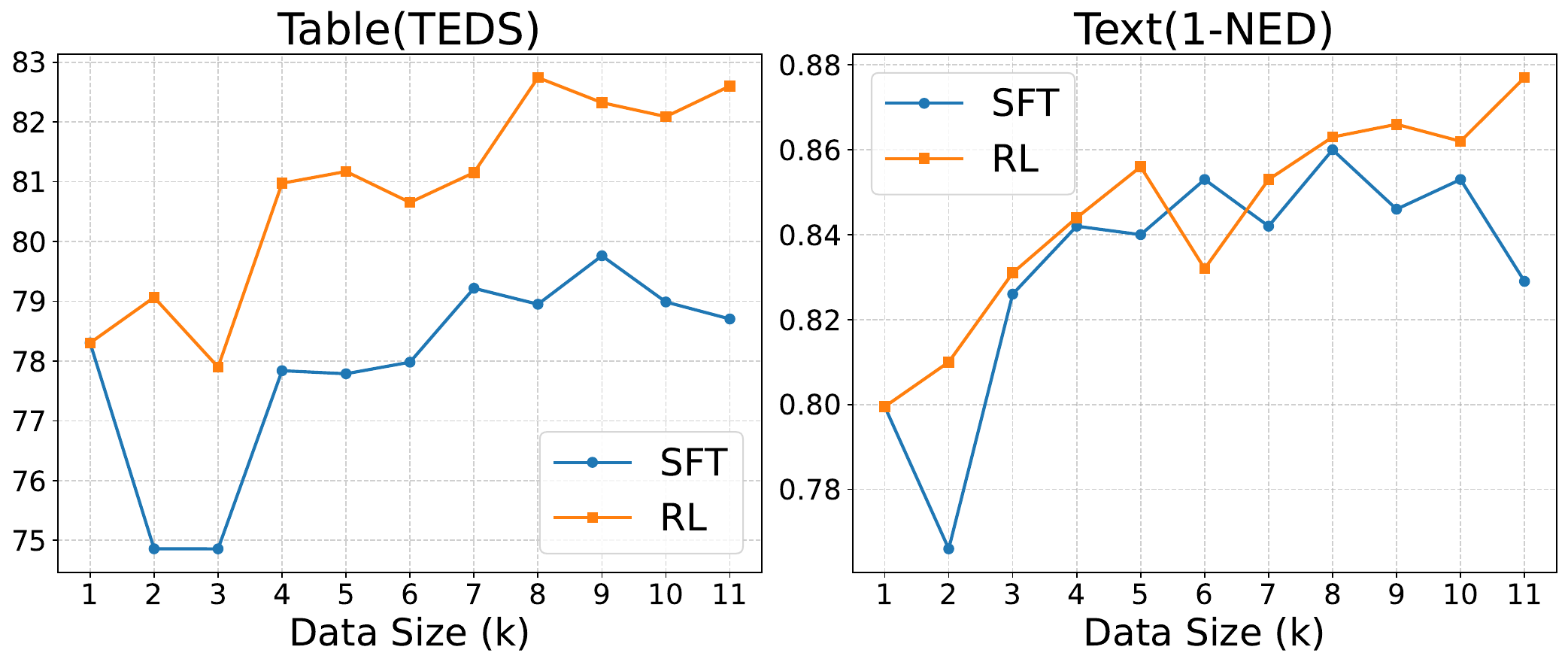}
        % \vspace{-2mm}
        % \caption{In-Distribution Task}
        \label{fig:metrics_indist}
    \end{subfigure}
    % 子图2
    \begin{subfigure}{0.49\textwidth}
        \centering
        \includegraphics[width=\linewidth]{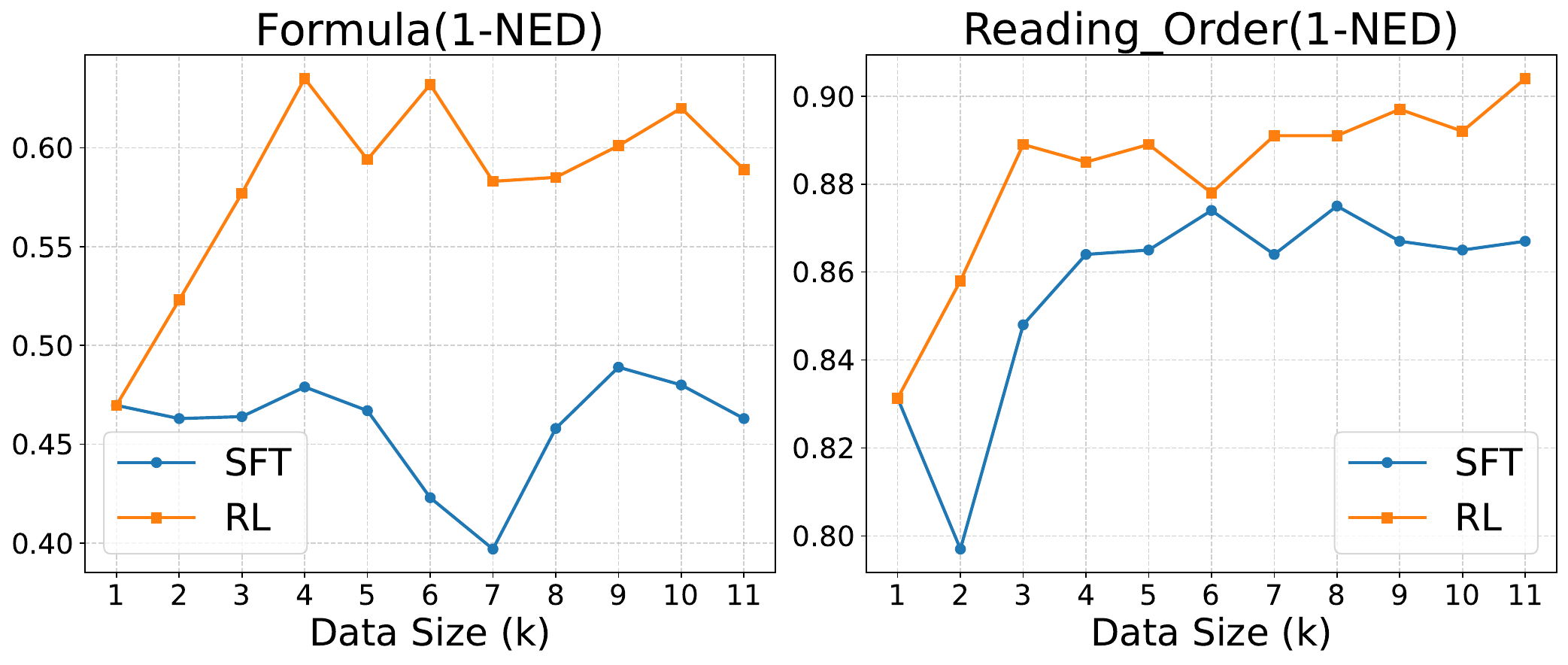}
        % \vspace{-2mm}
        % \caption{Out-of-Distribution Task}
        \label{fig:metrics_ood}
    \end{subfigure}
    \vspace{-5mm}
    \caption{Performance comparison of SFT and Layout-Aware RL on OmniDocBench sub-tasks.}
    \label{fig:metrics_comparison}
    \vspace{-2mm}
\end{figure}

\begin{figure}[h!]
    \centering
    \includegraphics[width=\textwidth]{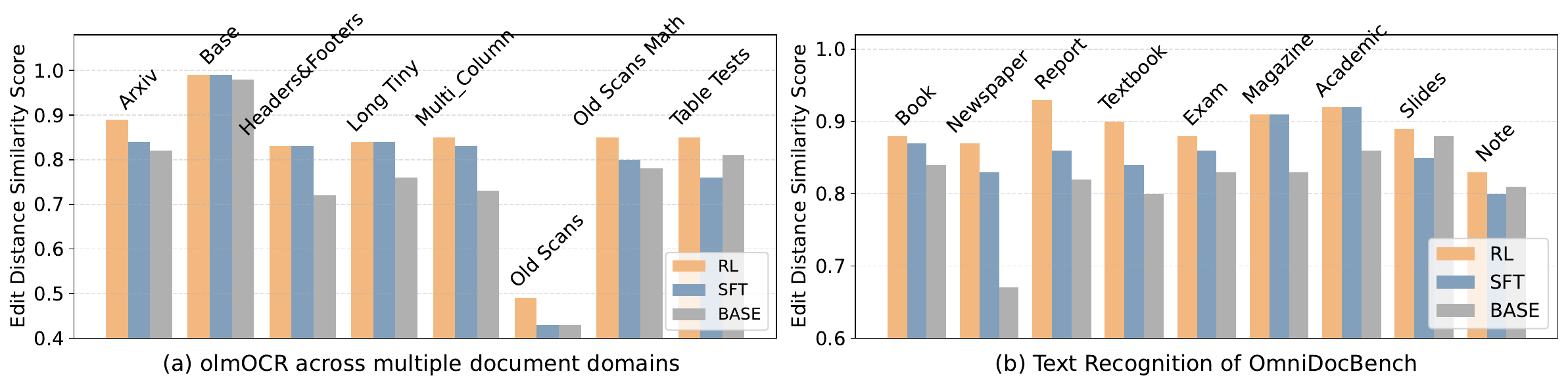}
    \vspace{-5mm}
    \caption{Comparison of model performance on different document parsering tasks.}
    \label{fig:olmocr-omnibench}
    \vspace{-3mm}
\end{figure}

\subsection{Further Analysis of LayoutRL}
\label{subsec:further_analysis}

We analyze the behavior of Layout-Aware RL compared with SFT across various document understanding settings. Our analysis focuses on three aspects: training stability, robustness across diverse document types, and generalization under distribution shifts. In particular, the third analysis examines generalization when the training data do not include the textbook and slides domains, which are used only for evaluation. This setup naturally forms OOD scenarios, allowing us to assess how well each model generalizes to unseen document types and layouts.

\paragraph{Training Stability Across Task Types.} As shown in Figure~\ref{fig:metrics_comparison}, we compare SFT and Layout-Aware RL on four OmniDocBench sub-tasks: table recognition, text recognition, formula parsing, and reading order prediction. Across all tasks, RL consistently achieves better performance, yielding higher scores on TEDS and lower errors on NED. More importantly, the RL curves exhibit smoother trajectories with stable improvements over training, while SFT shows large fluctuations and even performance regressions at certain stages. These results indicate that layout-aware rewards not only improve final accuracy but also enhance training stability throughout optimization.

\begin{figure}[t!]
    \centering
    \includegraphics[width=\textwidth]{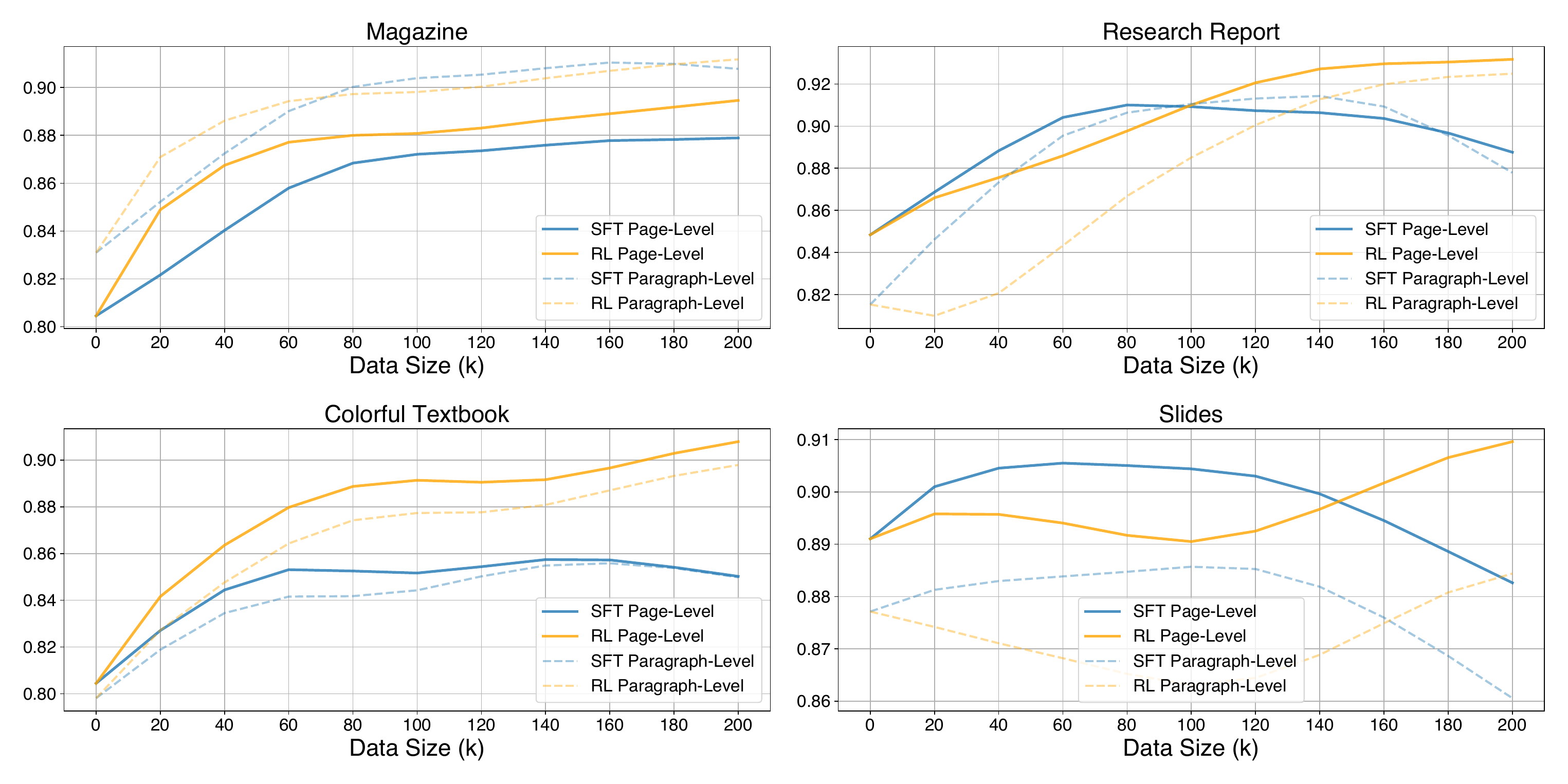}
    \vspace{-2mm}
    \caption{Generalization comparison of SFT and Layout-Aware RL. X-axis represents training steps. Y-axis represents 1-NED scores.}
    \label{fig:olmocr-omnibench222}
    \vspace{-3mm}
\end{figure}

\paragraph{Robustness Across Diverse Document Tasks.} Figure~\ref{fig:olmocr-omnibench} compares RL, SFT, and Zero-Shot (Base) across diverse document parsing tasks. On olmOCR (left), RL consistently achieves higher Levenshtein distance similarity score, especially on challenging cases like old scans and table tasks. At the same time, SFT offers moderate gains over Zero-Shot but remains behind RL. On OmniDocBench (right), RL also outperforms the other methods across most document types, showing notable improvements on books, reports, and academic texts. Overall, RL demonstrates greater robustness and better generalization in both structural parsing on olmOCR and text recognition on omnidocbench.

% \begin{figure}[ht]
%     \centering
%     \includegraphics[width=\textwidth]{picture/all_trend2.pdf}
%     \vspace{-2mm}
%     \caption{Generalization comparison of SFT and Layout-Aware RL. X-axis represents training steps. Y-axis represents 1-NED scores.}
%     \label{fig:olmocr-omnibench222}
%     \vspace{-3mm}
% \end{figure}

\paragraph{Analysis of Generalization} 
The Figure~\ref{fig:olmocr-omnibench222} compares document parsing performance across different training steps. The top two plots (magazine, research report) correspond to In-Distribution settings, where training and evaluation domains are aligned, while the bottom two plots (colorful textbook, slides) correspond to OOD settings, where evaluation involves unseen document types. In the in-distribution case, SFT achieves stable paragraph-level accuracy but its page-level performance tends to plateau, reflecting reliance on surface patterns. By contrast, RL continues to improve with data scale, achieving notable gains in page-level accuracy. In the OOD case, SFT performance degrades more severely, while RL maintains robustness and shows stronger improvements, highlighting its ability to capture global structural dependencies and generalize across distributions.

\section{Conclusion}
% In this work, we presented layoutRL, the first end-to-end reinforcement-learning framework that makes document parsers explicitly layout-aware by optimizing a multi-aspect reward—combining normalized edit-distance, paragraph-count accuracy, and reading-order preservation. To support this training paradigm, we released Infinity-Doc-55K, a 55K document corpus blending high-fidelity synthetic scanned document parsing data with expert-filtered real-world samples, providing the large-scale, precise supervision end-to-end models require. We implement the proposed method as a vision-language-model–based parser, Infinity-Parser. Powered by layoutRL, it achieves new state-of-the-art performance among end-to-end models on English and Chinese tasks, including OCR, table and formula extraction, and reading order detection. By releasing our code, dataset, and trained model, we hope to catalyze further advances in robust, reliable document understanding.

We introduced LayoutRL, an end-to-end reinforcement learning framework that explicitly incorporates layout awareness into document parsing through verifiable, multi-aspect rewards. To support this training, we built Infinity-Doc-400K, a large-scale dataset combining synthetic and real-world documents with diverse layouts, and trained Infinity-Parser, a VLM-based parser. Experiments on OmniDocBench, olmOCR-Bench, PubTabNet, and FinTabNet show that our approach achieves state-of-the-art performance across languages and document types, outperforming both specialized pipelines and general-purpose VLMs. Beyond accuracy, LayoutRL improves training stability and demonstrates robustness across diverse document tasks, highlighting reinforcement learning as a promising direction for robust and transferable document intelligence.



\bibliography{main}
\bibliographystyle{main}

\appendix

\newpage

% \section{Use of Large Language Models}

% We used a large language model (GPT4V) as an assistive tool during the preparation of this paper. 
% Specifically, it was employed for language polishing, grammar refinement, and providing alternative phrasings to improve readability. The model was not involved in generating research ideas, designing experiments, analyzing results, or drawing conclusions. All substantive contributions, including research conception, methodology, data analysis, and interpretation, were performed solely by the authors, who take full responsibility for the content.

% \section{Ethics Statement}

% This work does not involve human subjects, personal data, or sensitive user information. 
% All experiments are conducted on publicly available datasets or synthetic data generated via HTML rendering. 
% The newly introduced Infinity-Doc-400K dataset combines automatically generated synthetic documents with real-world samples that are pseudo-labeled through cross-model agreement and manually filtered to ensure quality. 
% We have taken care to remove potentially harmful or inappropriate content, and the dataset will be released strictly for research purposes under an academic license. 
% We believe our contributions do not pose risks of discrimination, bias, or privacy violations, and instead aim to advance the robustness and reliability of document parsing technologies for broad scientific and practical use.

\section{Reproducibility Statement}

We are committed to ensuring the reproducibility of our results. 
To this end, we will release the full Infinity-Doc-400K dataset, the implementation of LayoutRL with verifiable reward functions, and the pretrained Infinity-Parser model. 
Detailed training configurations, hyperparameters, and evaluation protocols are provided in the appendix and supplementary materials. 
Our experiments are conducted on widely used benchmarks, including OmniDocBench, olmOCR-Bench, PubTabNet, and FinTabNet, ensuring comparability with prior work. 
We will also provide scripts for preprocessing, training, and evaluation to facilitate reproducibility and further research by the community.

\section{Training Details}

\subsection{Training Context Length Distribution}

Our model was trained with a maximum context length of 8K tokens. To provide a clearer picture of the training data, we report detailed statistics of the context length distribution in Table~\ref{tab:context_length_summary} and Table~\ref{tab:context_length_distribution}. The average context length is 1,765 tokens, with a maximum of 31,147 tokens. More than 73\% of the samples fall within the [512, 4K) range. For sequences exceeding the 8K limit, we applied a left-truncation strategy to retain the semantically more relevant content at the end of the sequence.

\begin{table}[h!]
\centering

\begin{tabular}{lccccc}
\toprule
Metric & Min & Max & Average & Median & Std \\
\midrule
Value & 17 & 31,147 & 1,765 & 1,127 & 1,692 \\
\bottomrule
\end{tabular}
\caption{Summary statistics of training context length.}
\label{tab:context_length_summary}
\end{table}

\begin{table}[h!]
\centering
\resizebox{\textwidth}{!}{%
\begin{tabular}{lcccccccc}
\toprule
Context Length (tokens) & [1,256) & [256,512) & [512,1K) & [1K,2K) & [2K,4K) & [4K,8K) & [8K,16K) & $\geq$16K \\
\midrule
Frequency (count) & 25,125 & 40,680 & 119,041 & 102,757 & 70,554 & 39,780 & 2,482 & 63 \\
Distribution (\%) & 6.27 & 10.16 & 29.72 & 25.66 & 17.62 & 9.93 & 0.62 & 0.02 \\
\bottomrule
\end{tabular}}
\caption{Distribution of training samples across different context length intervals.}
\label{tab:context_length_distribution}
\end{table}

\section{Data Details}

\begin{figure}[ht]
  \centering
  \resizebox{1\linewidth}{!}{%
    \includegraphics{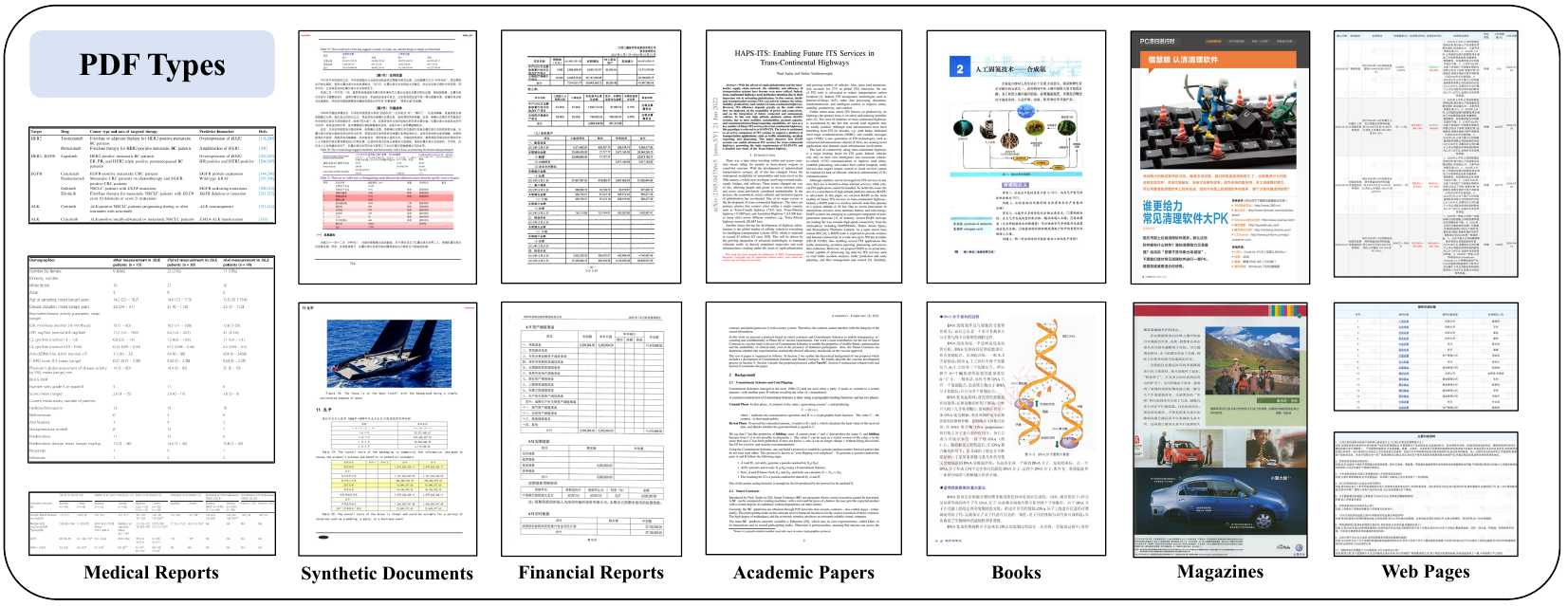}
  }
  \caption{This figure illustrates a diverse collection of PDF document types commonly encountered in Infinity-Doc-55K, grouped into seven categories: Medical Reports, Synthetic Documents, Financial Reports, Academic Papers, Books, Magazines, and Web Pages. }
  \label{fig:example1235}
\end{figure}

As illustrated in Figure~\ref{fig:example1235}, the dataset spans seven diverse document domains, making it one of the most richly annotated and structurally varied resources to date. Each domain is represented by two sample pages, highlighting the broad variability in layout design, content structure, and semantic density. For instance, Medical Reports typically contain structured tables with clinical measurements and diagnostic notes. Synthetic Documents are algorithmically generated to replicate real-world formats, providing layout diversity for training robust parsers. Financial Reports feature dense tables and formal accounting records, while Academic Papers often follow two-column layouts with references, equations, and figures. Books combine narrative content with visual illustrations, and Magazines blend images and stylized text for reader engagement. Finally, Web Pages, when saved as PDFs, preserve HTML-based structures that integrate tables, lists, and dynamic elements. This visual taxonomy exemplifies the structural and semantic diversity present in real-world documents, highlighting the core challenge faced by document AI systems: reliably parsing heterogeneous layouts and extracting structured information across a wide variety of formats.

Table \ref{tab:doc_breakdown_by_type} provides an overview of the document types included in the Infinity-Doc-400K dataset, detailing the composition across both real-world and synthetic sources. The real-world portion consists of 331K documents spanning six domains: financial reports, medical reports, academic papers, books, magazines, and web pages. These documents were collected from the web and annotated using a pseudo-labeling pipeline based on expert model agreement. While this approach enables large-scale data acquisition, the resulting label quality is relatively low due to occasional inconsistencies across models. Additionally, real-world data collection incurs a high cost, especially in terms of manual filtering, formatting normalization, and layout validation.

% In contrast, the synthetic subset includes 6.5k documents automatically generated using curated content from CC3M, Wikipedia, and the web, combined with predefined HTML templates. This pipeline allows for precise control over layout and annotation, resulting in high-quality labels at low cost. The combination of real and synthetic sources enables a balance between structural diversity, realism, and supervision quality, making Infinity-Doc-55K a comprehensive and scalable benchmark for training and evaluating document parsing models.

\begin{table}[h]
  \small
  \centering
  \setlength{\tabcolsep}{14pt} 
  \resizebox{1\linewidth}{!}{
    \begin{tabular}{l c c l c}
      \toprule
      \textbf{Data Source} & \textbf{Document Types} & \textbf{Size} & \textbf{Annotation Method} & \textbf{Cost} \\
      \midrule
      Real-World Doc & Financial Reports & 58.0K & Web + Pseudo-Label   & High \\
      Real-World Doc & Medical Reports   & 5.0K  & Web + Pseudo-Label   & High \\
      Real-World Doc & Academic Papers   & 71.7K  & Web + Pseudo-Label   & High \\
      Real-World Doc & Books             & 11.3K & Web + Pseudo-Label   & High \\
      Real-World Doc & Magazines         & 180.0K  & Web + Pseudo-Label   & High \\
      Real-World Doc & Web Pages         & 5.0K  & Web + Pseudo-Label   & High \\
      \midrule
      Synthetic      & Synthetic Documents & 69.0K & CC3M + Web + Wiki    & Low \\
      \bottomrule
    \end{tabular}
  }
  \vspace{3mm}
  \caption{Overview of document types in the Infinity-Doc-400K dataset, including data source, document type, annotation method, and collection cost.}
  \label{tab:doc_breakdown_by_type}
\end{table}

\section{Benchmarks Details}

\textbf{OmniDocBench~\citep{ouyang2024omnidocbench}} We conduct evaluation on OmniDocBench, a comprehensive benchmark that covers diverse document types and content modalities. To assess parsing performance across different structural elements, we employ two primary evaluation metrics: Normalized Edit Distance (NED), which measures the minimum edit operations required to transform one string into another normalized by the target string length, and Tree Edit Distance-based Similarity (TEDS), which captures structural similarities by comparing tree representations of HTML tables. These metrics are applied to different subtasks: NED is used to evaluate pure text, formula transcription, and reading order; TEDS combined with NED is used to evaluate both structural and content accuracy of table parsing.

% \textbf{Fox~\citep{liu2024focus_fox}}
% A multilingual benchmark for fine-grained, text-centric document understanding, covering 9 sub-tasks such as OCR, translation, summarization, layout analysis, and captioning. It includes 212 bilingual pages (112 English, 100 Chinese) with diverse layouts, evaluated using task-specific metrics like NED, F1, BLEU, METEOR, and ROUGE-L.

\textbf{PubTabNet~\citep{PubTabNet}}
A widely used benchmark for table recognition, containing 500,777 training and 9,115 validation images with diverse scientific table structures. Evaluation is conducted on the validation set.

\textbf{FinTabNet~\citep{zheng2021global}}
Focused on financial documents, this dataset includes 112,000 single-page scanned documents, with 92,000 cropped training images and 10,656 for testing. It features dense layouts and detailed annotations for both structure and content evaluation.

\paragraph{olmOCR-Bench~\citep{poznanski2025olmocr}} This is a benchmark developed to automatically and reliably evaluate document-level OCR performance across a wide range of tools. Unlike traditional evaluation metrics such as edit distance—which may penalize valid variations or fail to capture critical semantic errors—olmOCR-Bench focuses on verifying simple, unambiguous, and machine-checkable “facts” about each document page, similar to unit tests. For instance, it checks whether a specific sentence appears exactly in the OCR output. The benchmark operates directly on single-page PDFs to preserve digital metadata, which can be beneficial for certain OCR systems, and to maintain the integrity of the original document format. Designed for flexibility, olmOCR-Bench supports outputs in Markdown or plain text, allowing for seamless evaluation of both open-source and custom OCR pipelines.

\section{More Results}
% \section{Diverse Page Types Evaluation}
\paragraph{Diverse Page Types Evaluation}

To further investigate model behavior across diverse document types, we evaluated text recognition performance on nine distinct page categories. As shown in Table \ref{result_label2}, pipeline-based systems such as MinerU~\cite{MinerU} and Mathpix achieved strong results on structured formats like academic papers and financial reports. General-purpose vision-language models (VLMs) demonstrated better generalization on less formal page types, including presentation slides and handwritten notes. However, for challenging formats such as newspapers, most VLMs underperformed, while pipeline tools maintained relatively lower error rates. Notably, our proposed Infinity-Parser-7B achieved consistently low edit distances across all document types, outperforming both pipeline-based systems and general-purpose VLMs in overall accuracy. This highlights the robustness and adaptability of our reinforcement learning approach across diverse and complex document layouts.

\begin{table*}[h]
  % \centering
  \resizebox{1\textwidth}{!}{
    \begin{tabular}{l ccccccccc |c}
      \toprule
      \textbf{Models} & \textbf{Book} & \textbf{Slides} & \makecell{\textbf{Financial}\\\textbf{Report}} & \makecell{\textbf{Textbook}} & \makecell{\textbf{Exam}\\\textbf{Paper}} & \textbf{Magazine} & \makecell{\textbf{Academic}\\\textbf{Papers}} & \textbf{Notes} & \textbf{Newspaper} & \textbf{Overall $\downarrow$} \\
      \midrule
      \multicolumn{11}{l}{\textit{Based on Pipeline Tools}} \\
      MinerU & \textbf{0.055} & 0.124 & \textbf{0.033} & 0.102 & 0.159 &\textbf{ 0.072} & \textbf{0.025} & 0.984 & 0.171 & 0.206 \\
      Marker & 0.074 & 0.34 & 0.089 & 0.319 & 0.452 & 0.153 & 0.059 & 0.651 & 0.192 & 0.274 \\
      Mathpix & 0.131 & 0.22 & 0.202 & 0.216 & 0.278 & 0.147 & 0.091 & 0.634 & 0.69 & 0.3 \\
      \midrule
      \multicolumn{11}{l}{\textit{Based on Expert VLMs}} \\
      GOT-OCR & 0.111 & 0.222 & 0.067 & 0.132 & 0.204 & 0.198 & 0.179 & 0.388 & 0.771 & 0.267 \\
      Nougat & 0.734 & 0.958 & 1.000 & 0.820 & 0.930 & 0.83 & 0.214 & 0.991 & 0.871 & 0.806 \\
      \midrule
      \multicolumn{11}{l}{\textit{Based on General VLMs}} \\
      GPT-4o & 0.157 & 0.163 & 0.348 & 0.187 & 0.281 & 0.173 & 0.146 & 0.607 & 0.751 & 0.316 \\
      
      Qwen2-VL-72B & 0.096 & \textbf{0.061} & 0.047 & 0.149 & 0.195 & 0.071 & 0.085 & 0.168 & 0.676 & 0.179 \\
      InternVL2-76B & 0.216 & 0.098 & 0.162 & 0.184 & 0.247 & 0.150 & 0.419 & 0.226 & 0.903 & 0.3 \\
      
      % Qwen2.5-VL-7B & 0.662 & 0.341 & 0.238 & 0.743 & 0.421 & 0.574 & 0.671 & 0.342 & 0.820 & 0.536 \\

      % Qwen2.5-VL-7B & 0.178 & 0.125 & 0.178 & 0.256 & 0.178 & 0.203 & 0.222 & 0.249 & 0.738 & 0.259 \\

      Qwen2.5-VL-7B & 0.222 & 0.131 & 0.194 & 0.268 & 0.203 & 0.230 & 0.195 & 0.249 & 0.394 & 0.230 \\
      
      % InternVL3-8B & 0.514 & 0.488 & 0.331 & 0.385 & 0.276 & 0.438 & 0.493 & 0.277 & 0.758 & 0.445 \\

      InternVL3-8B & 0.311 & 0.233 & 0.320 & 0.222 & 0.238 & 0.157 & 0.438 & 0.268 & 0.726 & 0.328 \\
      
      \midrule
      \multicolumn{11}{l}{\textit{Based on Reinforcement Learning}} \\
      % PDF-SFT-7B & 0.216 & 0.098 & 0.162 & 0.184 & 0.247 & 0.150 & 0.419 & 0.226 & 0.903 & 0.3 \\
      Infinity-Parser-7B
        & 0.112 & 0.107 & 0.070 & \textbf{0.093} & \textbf{0.082} & 0.082 & 0.087 & \textbf{0.141} & \textbf{0.153}& \textbf{0.104} \\
      \bottomrule
    \end{tabular}
  }
  \caption{End-to-end text recognition performance on OmniDocBench: evaluation using edit distance across 9 PDF page types. We compare with Mathpix, MinerU~\citep{MinerU}, Marker~\citep{marker}, GOT-OCR~\citep{got2}, Nougat~\citep{Nougat}, GPT-4o~\citep{GPT3}, Qwen2-VL-72B~\citep{wang2024qwen2}, InternVL2-76B~\citep{InternVL}, Qwen2.5-VL-7B~\citep{bai2025qwen2.5}, InternVL3-8B~\citep{zhu2025internvl3}.}
  \label{result_label2}
  % \vspace{-3mm}
\end{table*}

Table~\ref{tab:table_single_attr} summarizes the performance of various models on the OmniDocBench table subset, evaluated along three dimensions: language diversity, table frame types, and special layout conditions. Notably, Infinity-Parser-7B achieves the best overall performance with an impressive score of 86.4, outperforming all other models across most individual metrics. It leads in nearly every category, including mixed-language settings (94.8), complex frame layouts (e.g., omission and three-line formats), and challenging special situations such as merged cells, formulas, and rotations. This demonstrates its strong generalization ability and robustness across diverse and noisy table formats.

\begin{table*}[h]
  \centering
  \renewcommand\tabcolsep{3pt}
  \resizebox{1\textwidth}{!}{
    \begin{tabular}{l|ccc|cccc|cccc|c}
    \toprule
    \multirow{2}{*}{\textbf{Model}} & \multicolumn{3}{c|}{\textbf{Language}} & \multicolumn{4}{c|}{\textbf{Table Frame Type}} & \multicolumn{4}{c|}{\textbf{Special Situation}} & \multirow{2}{*}{\textbf{Overall} $\uparrow$ }  \\
    & \textit{EN} & \textit{ZH} & \textit{Mixed}   & \textit{Full} & \textit{Omission} & \textit{Three} & \textit{Zero} & \textit{Merge Cell}(+/-) & \textit{Formula}(+/-) & \textit{\makecell{Colorful}}(+/-) & \textit{Rotate}(+/-) \\
    \midrule
    PaddleOCR~\citep{li2022ppocrv3attemptsimprovementultra} & 76.8 & 71.8 & 80.1 & 67.9 & 74.3 & 81.1 & 74.5 & 70.6/75.2 & 71.3/74.1 & 72.7/74.0 & 23.3/74.6 & 73.6 \\
    RapidTable~\citep{RapidTable2023} & 80.0 & 83.2 & 91.2 & 83.0 & 79.7 & 83.4 & 78.4 & 77.1/85.4 & 76.7/83.9 & 77.6/84.9 & 25.2/83.7 & 82.5 \\
    StructEqTable~\citep{StructEqTable2024} & 72.8 & 75.9 & 83.4 & 72.9 & 76.2 & 76.9 & 88.0 & 64.5/81.0 & 69.2/76.6 & 72.8/76.4 & 30.5/76.2 & 75.8 \\
    GOT-OCR~\citep{got2} & 72.2 & 75.5 & 85.4 & 73.1 & 72.7 & 78.2 & 75.7 & 65.0/80.2 & 64.3/77.3 & 70.8/76.9 & 8.5/76.3 & 74.9 \\
    Qwen2-VL-7B~\citep{wang2024qwen2} & 70.2 & 70.7 & 82.4 & 70.2 & 62.8 & 74.5 & 80.3 & 60.8/76.5 & 63.8/72.6 & 71.4/70.8 & 20.0/72.1 & 71.0 \\
    InternVL2-8B~\citep{InternVL} & 70.9 & 71.5 & 77.4 & 69.5 & 69.2 & 74.8 & 75.8 & 58.7/78.4 & 62.4/73.6 & 68.2/73.1 & 20.4/72.6 & 71.5 \\
    Qwen2.5-VL-7B~\citep{wang2024qwen2} & 87.4 & 80.7 & 93.5 & 86.4 & 85.1 & 84.1 & 88.7 & 77.5/89.8 & 82.1/87.2 & 77.1/87.5 & 56.5/86.0 & 85.5 \\
    InternVL3-8B~\citep{zhu2025internvl3} & 79.5 & 86.0 & 91.7 & 85.5 & 80.7 & 83.9 & 85.9 & 71.9/90.9 & 74.0/86.7 & 82.1/85.3 & 12.6/85.5 & 84.3 \\
    \midrule
    \textbf{Infinity-Parser-7B} & 84.7 & \textbf{86.7} & \textbf{94.8} & \ 85.5 & \textbf{86.5 }& \textbf{87.4} & \textbf{89.4} & \textbf{78.6/90.7} & 81.9/ \textbf{87.5} & \textbf{83.2/88.0} & \textbf{68.8/86.7} & \textbf{86.4} \\
    \bottomrule
    \end{tabular}%
  }
  \caption{Component-level Table Recognition evaluation on OmniDocBench table subset. \textit{(+/-)} means \textit{with/without} special situation. }
  \label{tab:table_single_attr}%
  % \vspace{-5mm}
\end{table*}

\section{Prompt Strategy for Parsing Tasks.}

\textbf{Prompt Template} summarizes the prompt designs for two key parsing tasks: document parsing and table parsing. For document parsing, the prompts instruct the model to recognize visual regions and convert their contents into structured Markdown. This design ensures consistent region-level extraction across documents with diverse layouts.

For table parsing, although the prompts are phrased differently, they share the same objective: transforming table content from images into HTML. This diversity encourages the model to generalize across variations in phrasing and reduces overfitting to a single instruction template. Notably, HTML is used here to match the evaluation format, but the resulting outputs can be easily converted to Markdown if needed for downstream use.

\begin{figure}[t]
  \centering
  \includegraphics[width=\linewidth]{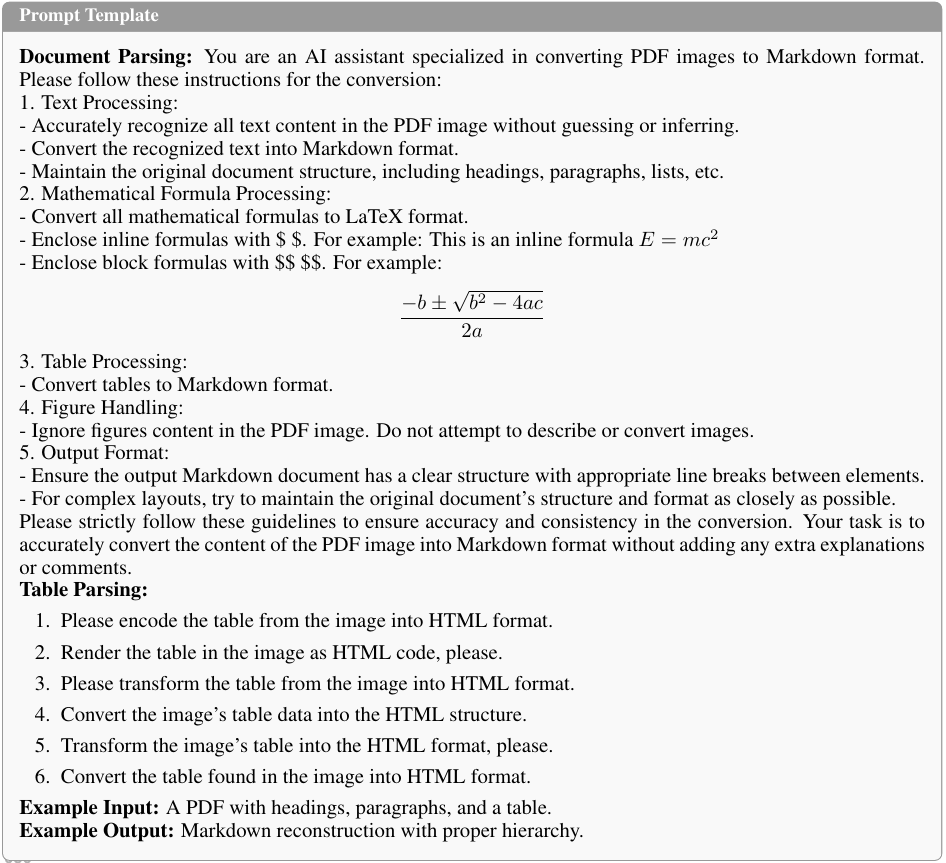} % 改成你的路径
  % \caption{Your caption here.}
  \label{fig:myplot}
\end{figure}

\section{Case Analysis}

\begin{figure}[htbp]
  \centering
  \resizebox{1\linewidth}{!}{%
    \includegraphics{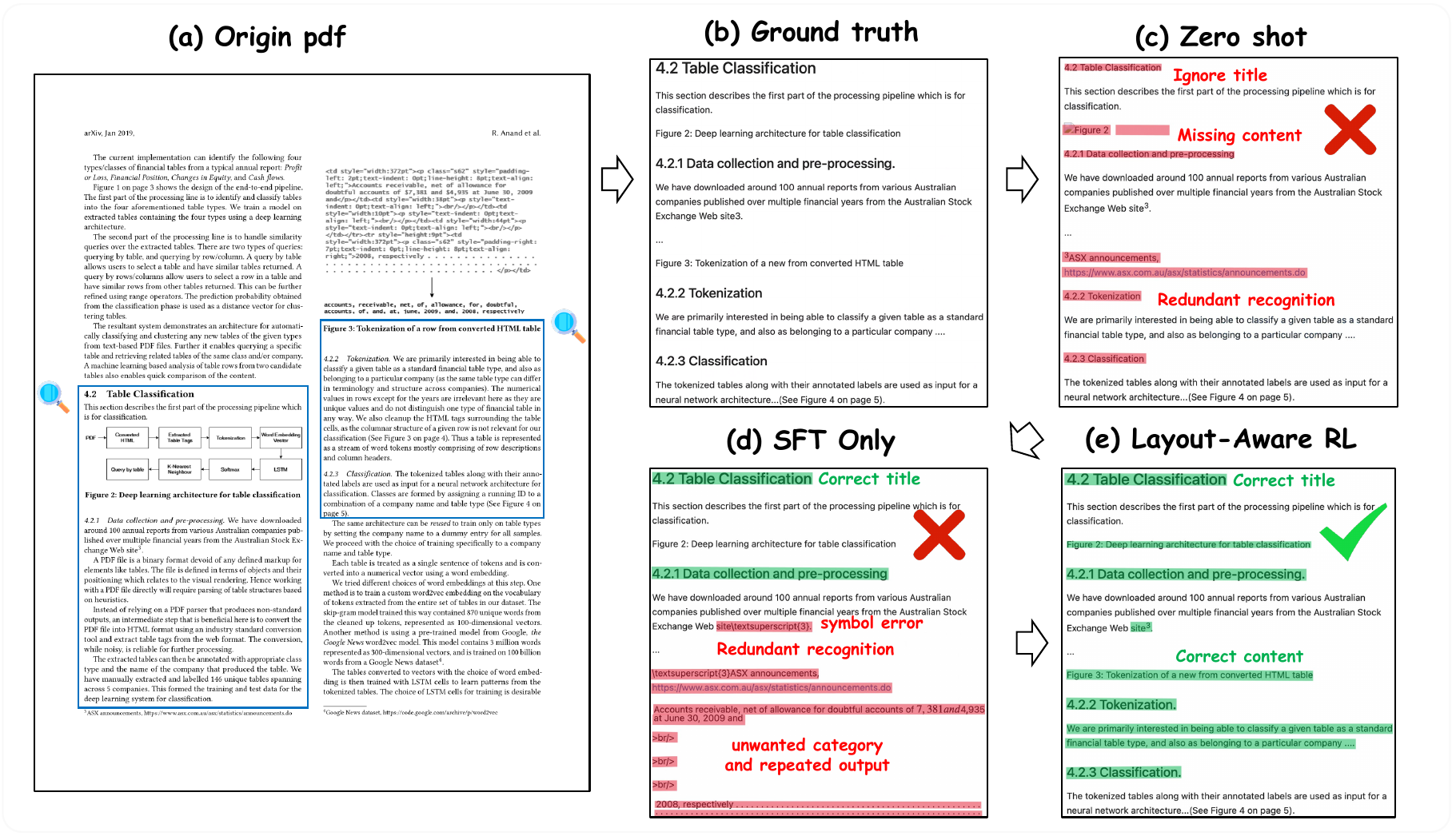}
  }
  \caption{Comparison of four Markdown generation results on a single case, illustrating progressive improvement from direct inference to full reward integration.}
  \label{fig:example1223}
\end{figure}

Figure~\ref{fig:example1223} illustrates a progressive improvement in Markdown generation quality across different training strategies. The zero-shot model fails to capture key structural elements, omitting titles and producing redundant or incomplete content. With SFT, the model better identifies section headers and general layout but still suffers from symbol-level errors and repeated outputs. In contrast, the layout-aware RL model demonstrates the most accurate and coherent result, successfully preserving the document hierarchy and eliminating redundancy. This highlights the effectiveness of layout-aware rewards in guiding the model toward semantically and structurally faithful document parsing.

\begin{figure}[htbp]
  \centering
  \resizebox{1\linewidth}{!}{%
    \includegraphics{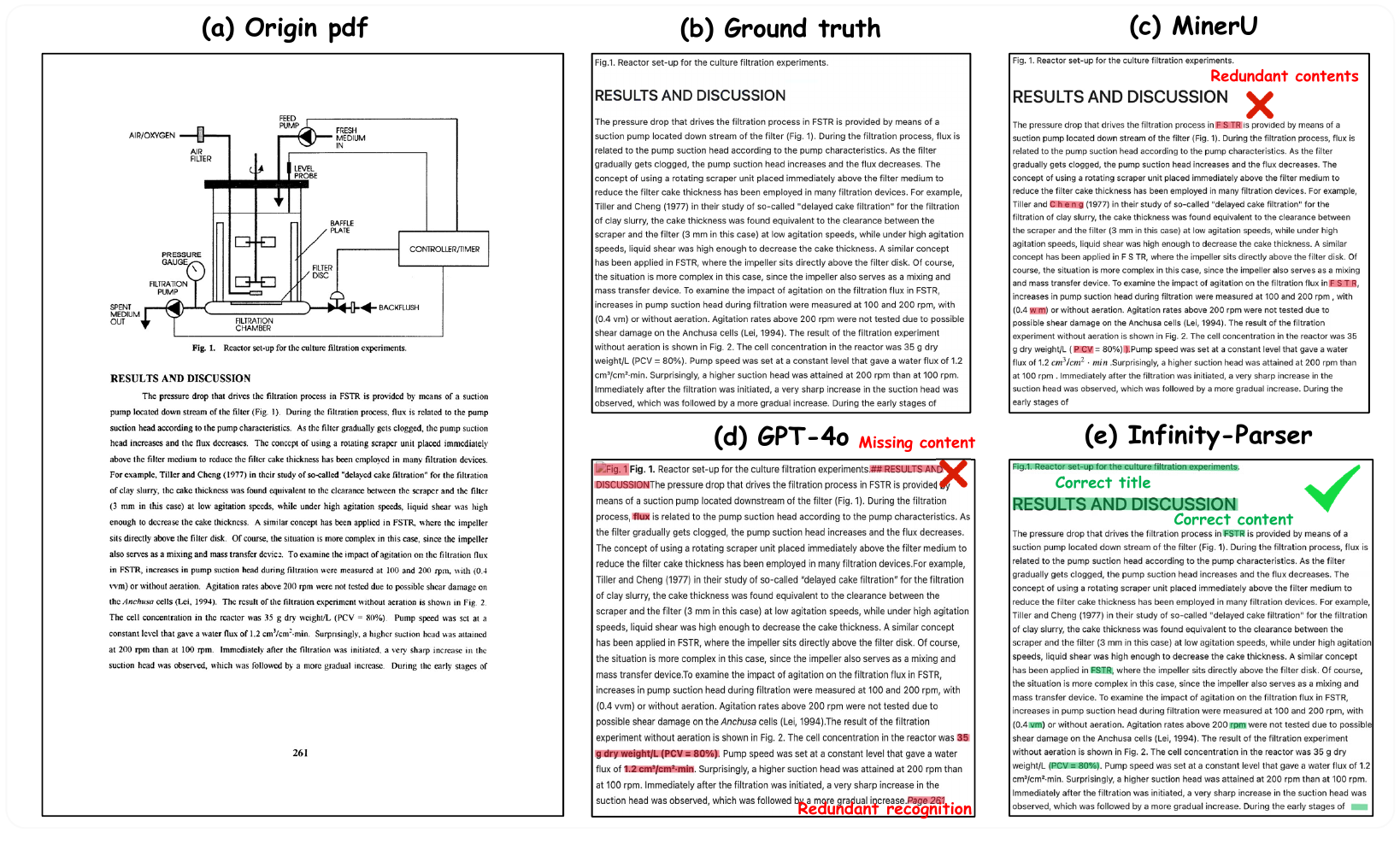}
  }
  \caption{Comparison of Markdown extraction from academic literature using different models. The figure shows the original PDF (a), ground truth annotations (b), and extraction results from three models: MinerU, GPT-4o, and Infinity-Parser (c–e). Infinity-Parser produces the most accurate output, correctly identifying titles and content while avoiding redundancy and omissions.}
  \label{fig:example123_academic}
\end{figure}

\begin{figure}[htbp]
  \centering
  \resizebox{1\linewidth}{!}{%
    \includegraphics{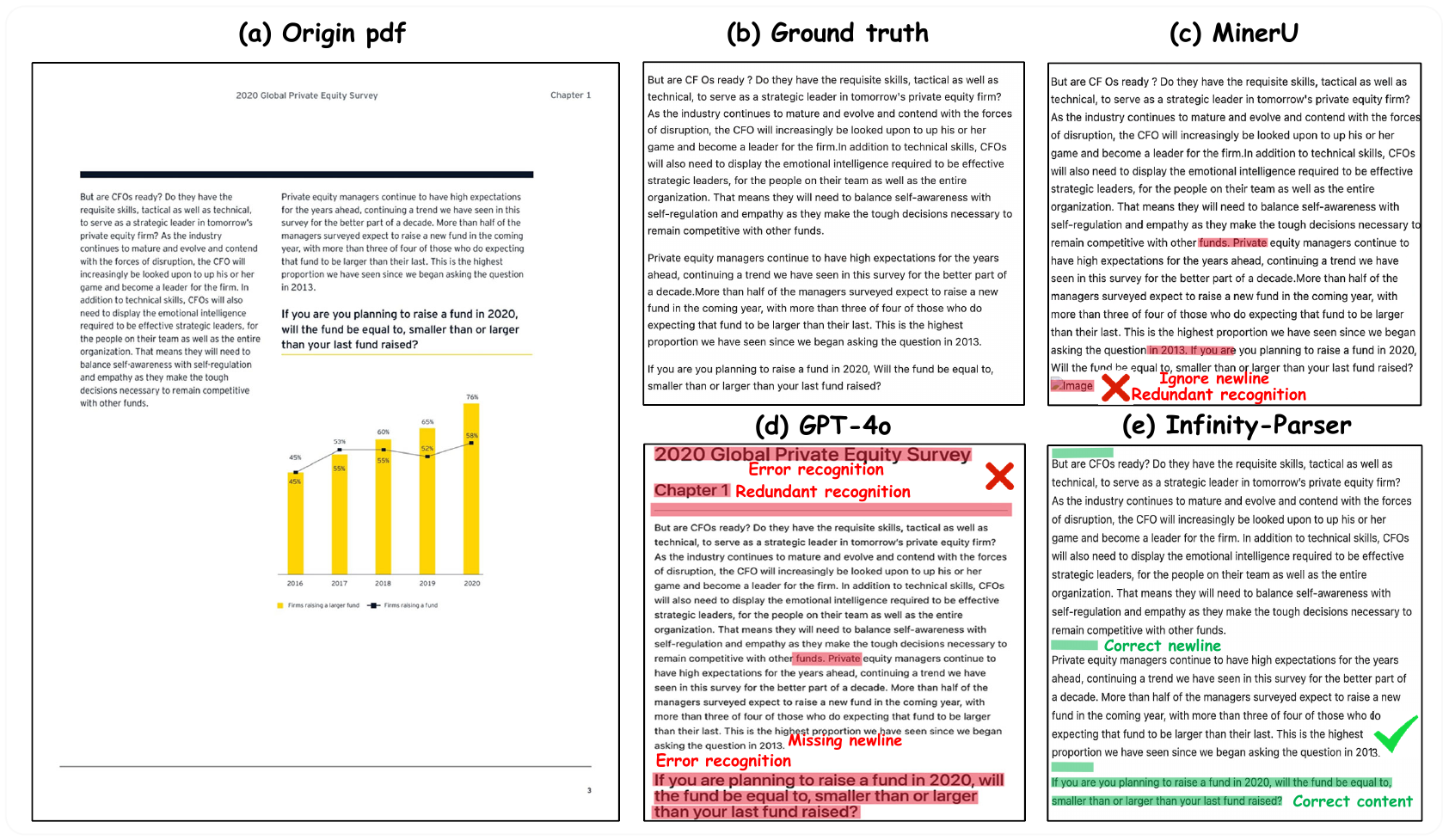}
  }
  \caption{Comparison of Markdown extraction from a book-style PDF using different models. This figure displays the original document (a), human-annotated ground truth (b), and the extraction results from MinerU, GPT-4o, and Infinity-Parser (c–e). GPT-4o introduces multiple errors, such as incorrect headings, redundancy, and missing lines. MinerU retains unnecessary line breaks and repeated text. In contrast, Infinity-Parser correctly identifies the content structure, including titles and paragraphs, producing clean and accurate output.}
  \label{fig:example123_book}
\end{figure}

\begin{figure}[htbp]
  \centering
  \resizebox{1\linewidth}{!}{%
    \includegraphics{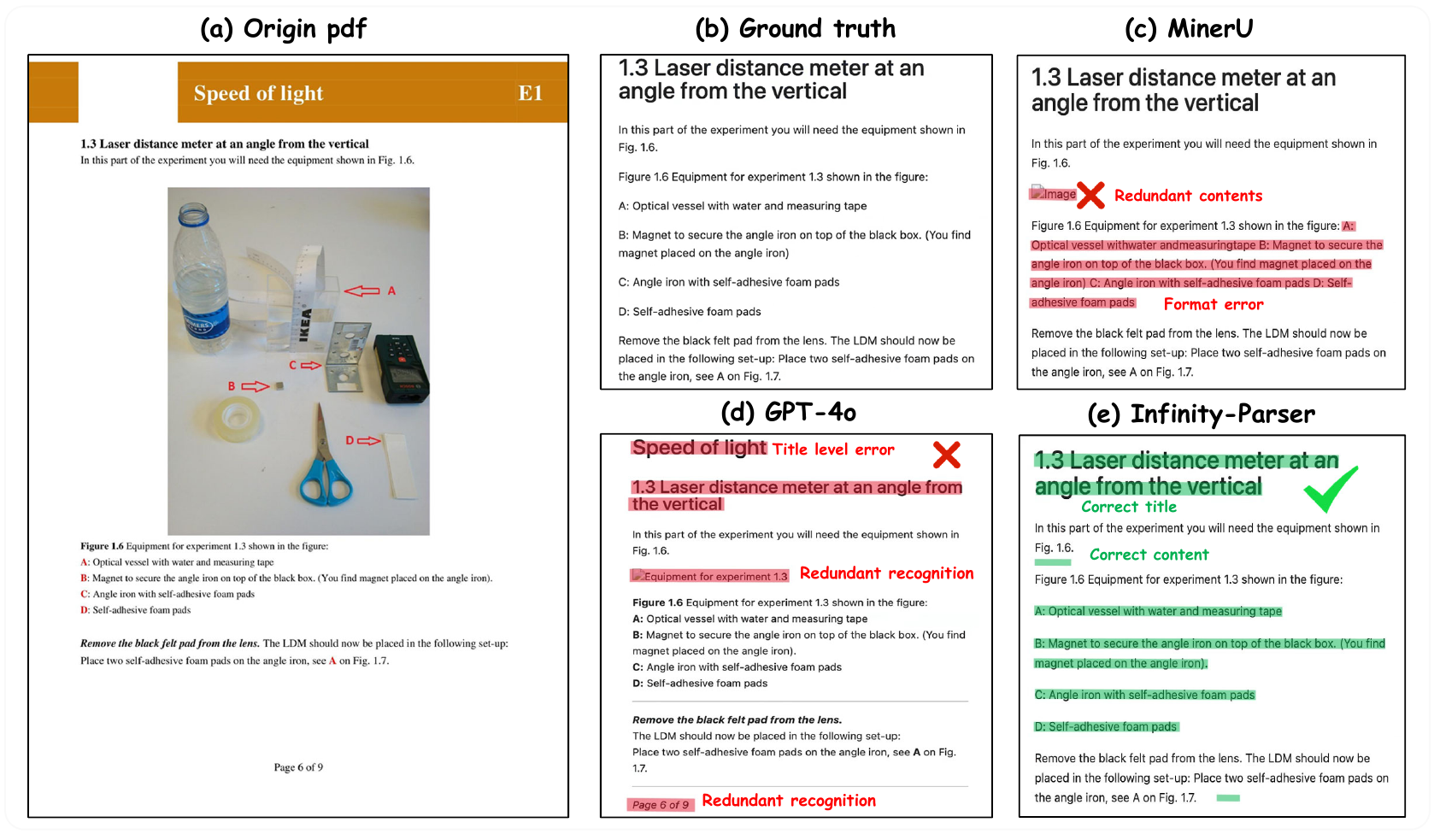}
  }
  \caption{Comparison of Markdown extraction from an exam-style PDF. The figure presents the original exam document (a), ground truth annotations (b), and the extraction results from MinerU, GPT-4o, and Infinity-Parser (c–e). GPT-4o and MinerU both introduce redundant text and formatting errors, such as incorrect title levels and misplaced content. In contrast, Infinity-Parser accurately captures the heading hierarchy and structured list format, faithfully reproducing the content as intended in the original layout.}
  \label{fig:example123_exam}
\end{figure}

\begin{figure}[htbp]
  \centering
  \resizebox{1\linewidth}{!}{%
    \includegraphics{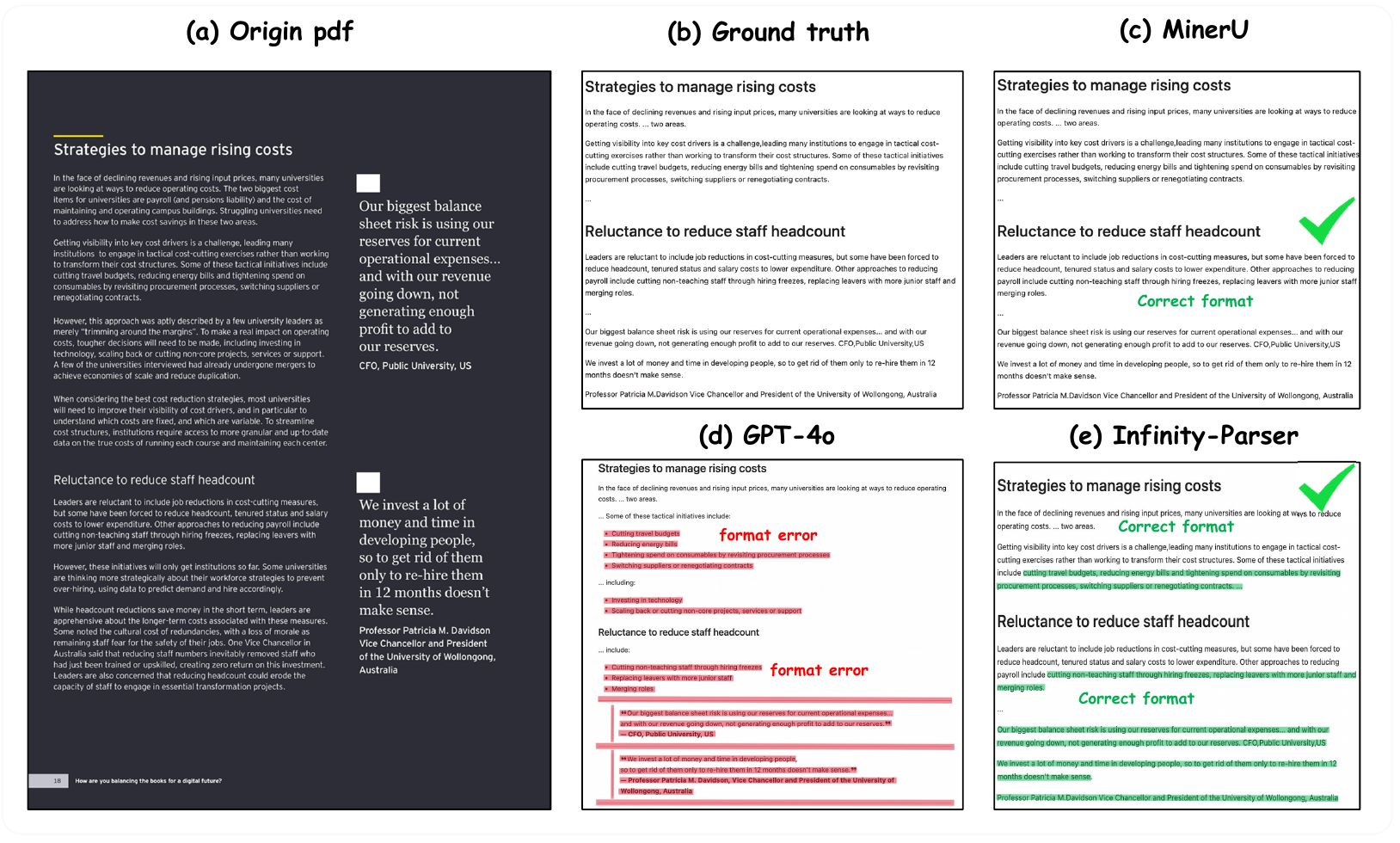}
  }
  \caption{Comparison of Markdown extraction from a magazine-style PDF. The figure shows the original visually-rich page (a), ground truth annotations (b), and results from MinerU, GPT-4o, and Infinity-Parser (c–e). Due to the complex layout and dark background, GPT-4o suffers from significant formatting errors. Infinity-Parser accurately preserves the structural hierarchy and formatting, demonstrating robustness in handling stylized layouts.}
  \label{fig:example123_magazine}
\end{figure}

\begin{figure}[htbp]
  \centering
  \resizebox{1\linewidth}{!}{%
    \includegraphics{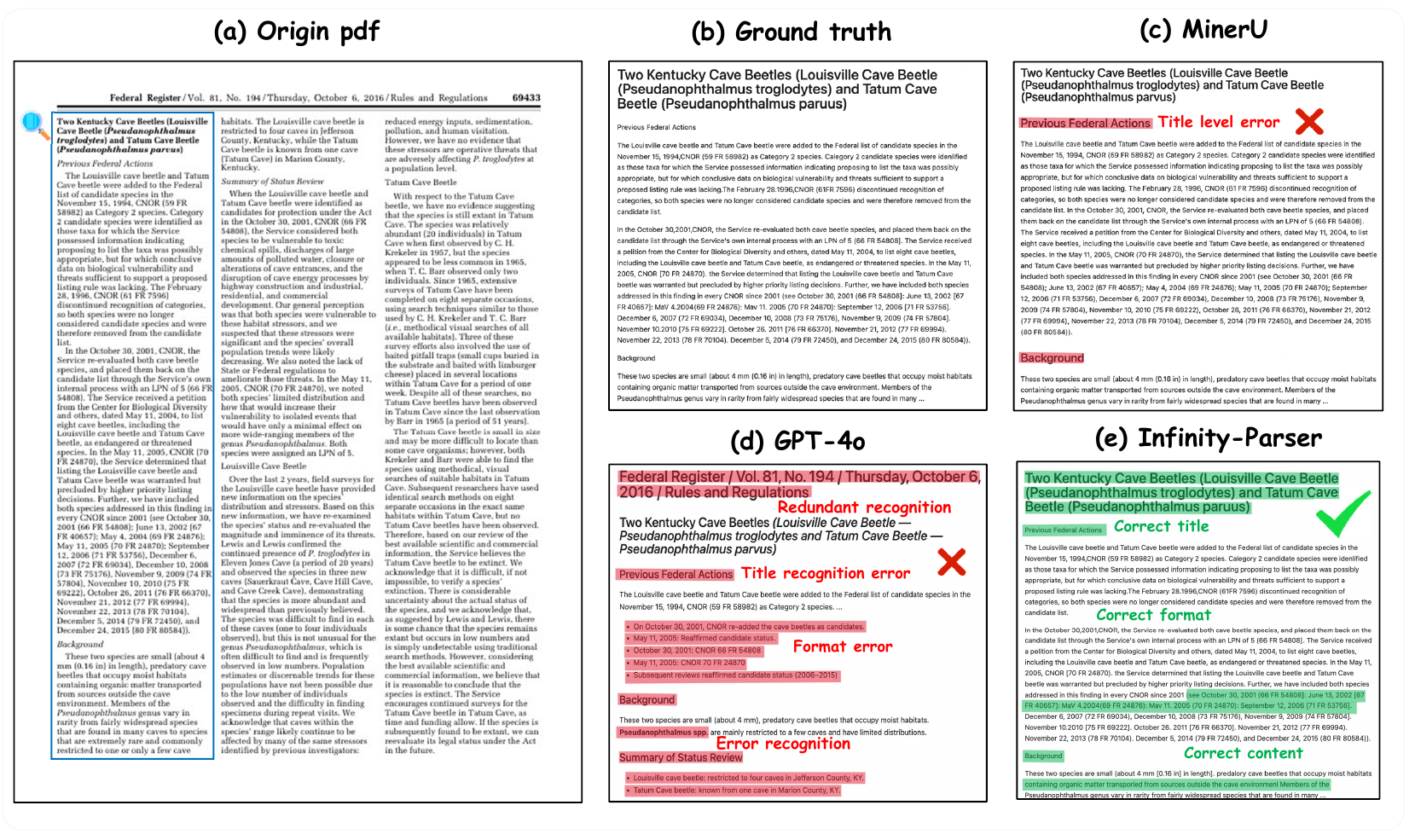}
  }
  \caption{ Comparison of Markdown extraction from a newspaper-style PDF. This figure presents the original densely formatted page (a), ground truth annotations (b), and the results from MinerU, GPT-4o, and Infinity-Parser (c–e). Due to the complex multi-column layout and title hierarchy, both MinerU and GPT-4o exhibit issues such as incorrect title levels, redundant content, and format errors. In contrast, Infinity-Parser accurately identifies the main title, maintains structural formatting, and preserves content integrity, demonstrating strong layout understanding in challenging document types.}
  \label{fig:example123_newspaper}
\end{figure}

\begin{figure}[htbp]
  \centering
  \resizebox{1\linewidth}{!}{%
    \includegraphics{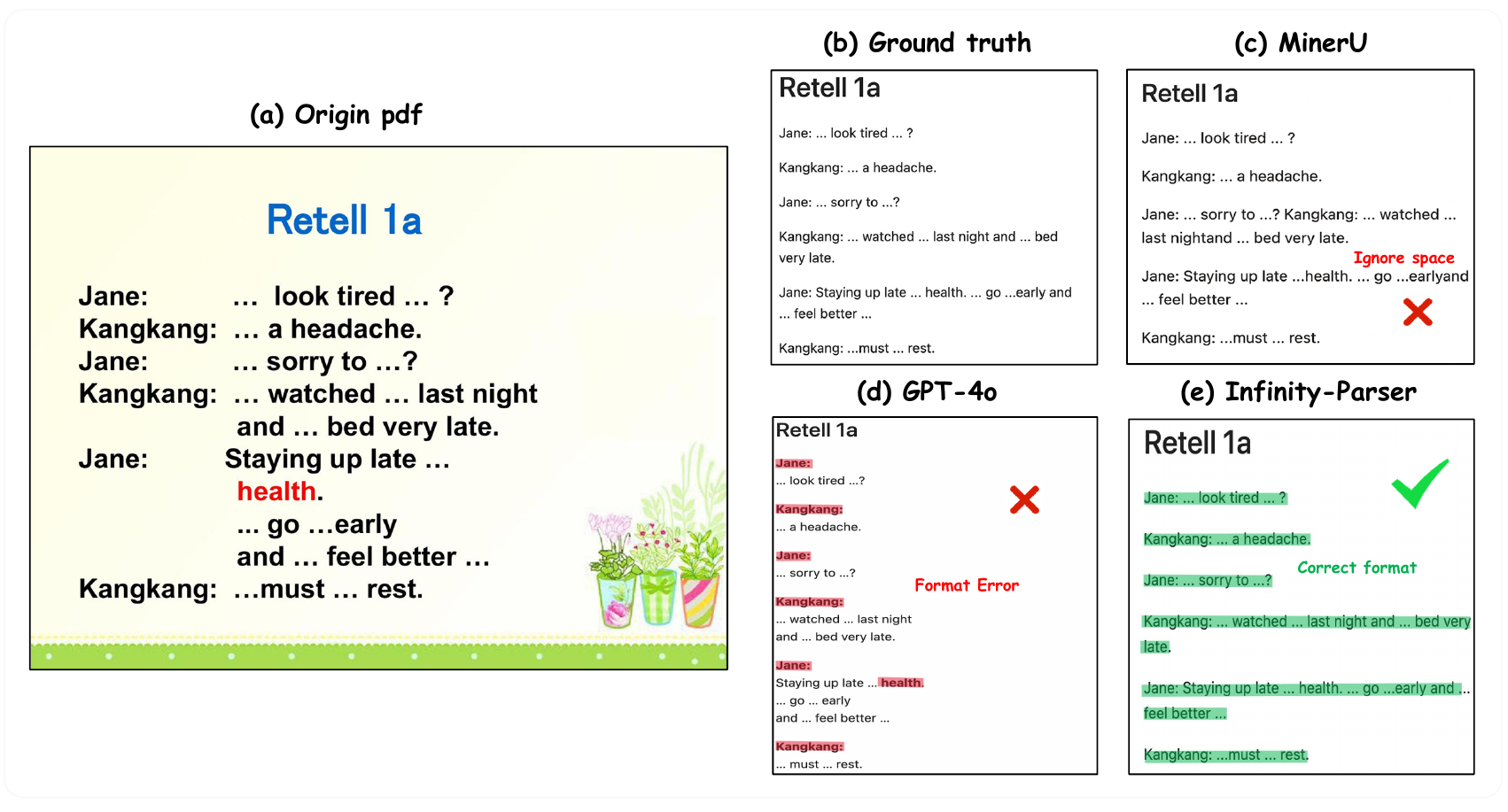}
  }
  \caption{Comparison of Markdown extraction from a PowerPoint-style PDF slide. The figure includes the original slide (a), human-annotated ground truth (b), and outputs from MinerU, GPT-4o, and Infinity-Parser (c–e). MinerU and GPT-4o both struggle with layout fidelity, introducing spacing and formatting errors. In contrast, Infinity-Parser preserves the dialogue structure and formatting accurately, demonstrating its capability to handle informal, visually decorated slides effectively.}
  \label{fig:example123_ppt}
\end{figure}

Infinity-Parser exhibits consistent improvements across a wide spectrum of document types, including academic papers, books, colorful textbooks, exam papers, magazines, government notices, newspaper articles, and PowerPoint-style slides. These gains are reflected in structural parsing, title and content recognition, formatting accuracy, and robustness to diverse visual layouts. As shown in Figures~\ref{fig:example123_academic} through~\ref{fig:example123_ppt}, we provide detailed visual comparisons with existing models, where Infinity-Parser consistently achieves superior results. These findings underscore the effectiveness and generalizability of our layout-aware RL approach across complex, real-world PDF formats.

\newpage

\end{document}